# *BioEVAL*: A global, multi-institutional benchmark of large language and multimodal models for bioengineering

Shun Ye[1,28,‡], Vinny Chandran Suja[2,3,‡], Chenlong Li[1,4], Chongming Jiang[5], Reza Zamani[5], Xiang Li[1,4], Christopher Bain[6,7], Yuqi Zhou[8,#], Walker Peterson[8,#], Huidong Wang[8,#], Chenglang Hu[8,#], Jongchan Park[1,#], Xiao Cheng[4,#], Benjamin Swedlund[9,#], Sandra Murillo[9,#], Anjali Sivanandan[1,#], Shiyu Sun[10,#], Liang Lanfeng[11,#], Mohammad Tariqul Islam[12,#], Baju C. Joy[12,#], Ishaq N. Khan[12,#], Sreedhar S. Kumar[13,#], Gabriel Mercado-Vásquez[14,15,#], James V. Vizzard[14,15,#], Jonathan M. Matthews[14,16,#], Helen Huang[17,#], Xiaolu Guo[17,#], Ethan Nicklow[18,#], Guorui Chen[1,#], Ryan A. Neff[19,#], Surjendu Maity[20,#], Hyeonjin Park[21,22,#], Han-ho Joo[21,22,#], Katherine Dong[23,24,#], Yuyan Cai[25,#], Weihang Huang[1,26], Yichen Zou[27,28], Rui Yan[27,28], Raphael Figueroa[28], Artem Goncharov[27,28], Bella Rose Schremmer[1,28], Lian Elsa Linton[28], Keisuke Goda[1,8,29,30,31,32#], Liang Gao[1,#], Ke Cheng[4,#], Leonardo Morsut[9,33#], Jennifer L. Wilson[1,#], Jianping Fu[10,34,35#], Lim Chwee Teck[11,36,37#], Deblina Sarkar[12,#], Andreas Hierlemann[13,#], Savaş Tay[14,15,#], Alexander Hoffmann[17,#], Donald Richieri Griffin[18,38,#], Jun Chen[1,#], Shana O. Kelley[19,39,40,#], Shyni Varghese[20,41,42,43#], Jinwoo Cheon[21,22,44,45,#], Wilbur A. Lam[6,7,46,47,#], James J. Moon[23,24,48,49,50,#], Wilson W. Wong[25,#], Samir Mitragotri[2,3,#], and Dino Di Carlo[1,28,51,52,53,*]

[1] Department of Bioengineering, University of California, Los Angeles, CA 90095, USA
[2] Harvard John A. Paulson School of Engineering and Applied Sciences, Harvard University, Allston, MA 02134, USA
[3] Wyss Institute for Biologically Inspired Engineering, Boston, MA 02215, USA
[4] Department of Biomedical Engineering, Columbia University, New York, NY 10027, USA
[5] Terasaki Institute for Biomedical Innovation, Woodland Hills, CA 91367, USA
[6] The Wallace H. Coulter Department of Biomedical Engineering, Georgia Institute of Technology, Atlanta, GA 30332, USA
[7] Parker H. Petit Institute of Bioengineering and Bioscience, Georgia Institute of Technology, Atlanta, GA 30332, USA
[8] Department of Chemistry, The University of Tokyo, Tokyo 113-0033, Japan
[9] Eli and Edythe Broad CIRM Center for Regenerative Medicine and Stem Cell Research, Keck School of Medicine, University of Southern California, Los Angeles, CA 90033, USA
[10] Department of Mechanical Engineering, University of Michigan, Ann Arbor, MI 48109, USA
[11] Mechanobiology Institute, National University of Singapore, Singapore 117411, Singapore
[12] Nano-Cybernetic Biotrek, Media Lab, Massachusetts Institute of Technology, Cambridge, MA 02139, USA
[13] Department of Biosystems Science and Engineering, ETH Zurich, CH-4056 Basel, Switzerland
[14] Pritzker School of Molecular Engineering, University of Chicago, Chicago, IL 60637, USA.
[15] Institute for Genomics and Systems Biology, University of Chicago, Chicago, IL, USA.
[16] Department of Medicine, Biological Sciences Division, University of Chicago, Chicago, IL 60637, USA
[17] Institute for Quantitative and Computational Biosciences, University of California, Los Angeles, CA 90095, USA
[18] Department of Biomedical Engineering, University of Virginia, Charlottesville, VA 22903, USA
[19] Department of Biomedical Engineering, McCormick School of Engineering, Northwestern University, Evanston, IL 60208, USA
[20] Department of Orthopaedic Surgery, Duke University School of Medicine, Durham, NC 27705, USA
[21] Center for Nanomedicine, Institute for Basic Science (IBS), Seoul 03722, Republic of Korea
[22] Department of Nano Biomedical Engineering (NanoBME), Institute for Advanced Science, Yonsei University, Seoul 03722, Republic of Korea
[23] Department of Pharmaceutical Sciences, University of Michigan, Ann Arbor, MI 48109, USA
[24] Biointerfaces Institute, University of Michigan, Ann Arbor, MI 48109, USA
[25] Department of Biomedical Engineering and Biological Design Center, Boston University, Boston, MA 02215, USA

[26] Department of Linguistics and Communication, University of Birmingham, Edgbaston, Birmingham, B15 2TT, UK
[27] Department of Electrical & Computer Engineering, University of California, Los Angeles, CA 90095, USA
[28] California NanoSystems Institute (CNSI), Los Angeles, CA 90095, USA
[29] SiRIUS Institute of Medical Research, Tohoku University, Miyagi 980-0872, Japan
[30] Graduate School of Medicine, Tohoku University, Miyagi 980-0872, Japan
[31] International Centre for Synchrotron Radiation Innovation Smart, Tohoku University, Miyagi 980-8577, Japan
[32] Healthspan Research Center, Tohoku University, Miyagi 980-0872, Japan
[33] Department of Biomedical Engineering, Viterbi School of Engineering, University of Southern California, Los Angeles, CA 90033, USA
[34] Department of Cell & Developmental Biology, University of Michigan Medical School, Ann Arbor, MI 48109, USA
[35] Department of Biomedical Engineering, University of Michigan, Ann Arbor, MI 48109, USA
[36] Department of Biomedical Engineering, National University of Singapore, Singapore 117583, Singapore
[37] Institute for Health Innovation and Technology (iHealthtech), National University of Singapore, Singapore 117599, Singapore
[38] Department of Chemical Engineering, University of Virginia, Charlottesville, VA 22903, USA
[39] Department of Chemistry, Weinberg College of Arts & Sciences, Northwestern University, Evanston, IL 60208, USA
[40] Biohub, Chicago, IL 60642, USA
[41] Department of Orthopaedic Surgery, University of California, Los Angeles, CA 90095, USA
[42] Department of Biomedical Engineering, Duke University, Durham, NC 27708, USA
[43] Department of Mechanical Engineering and Materials Science, Duke University, Durham, NC 27705, USA
[44] Department of Chemistry, Yonsei University, Seoul 03722, Republic of Korea
[45] Max Planck-Yonsei IBS Center for Nanomedicine Deep Tissue Control, Seoul 03722, Republic of Korea
[46] Department of Pediatrics, Emory University School of Medicine, Atlanta, GA 30322, USA
[47] Aflac Cancer and Blood Disorders Center of Children's Healthcare of Atlanta, Atlanta, GA 30322, USA
[48] Department of Chemical Engineering, University of Michigan, Ann Arbor, MI 48109, USA
[49] Department of Biomedical Engineering, University of Michigan, Ann Arbor, MI 48109, USA
[50] Rogel Cancer Center, University of Michigan, Ann Arbor, MI 48109, USA
[51] EXPERT-GRI Office and Graduate School of Science, The University of Tokyo, Tokyo 113-0033, Japan
[52] Jonsson Comprehensive Cancer Center, University of California, Los Angeles, CA 90095, USA
[53] Department of Mechanical and Aerospace Engineering, University of California, Los Angeles, CA 90095, USA

[‡] These authors contributed equally. Emails: shunye@ucla.edu (S.Y.); vinny@seas.harvard.edu (V.C.S.)
[#] These ***BioEVAL*** consortium authors contributed equally.

[*] Corresponding author. Email: dicarlo@ucla.edu (D.D.)

## I. Abstract

Large Language Models (LLMs) have demonstrated historic breakthroughs in general reasoning with early successes in biomedical science. However, existing LLM benchmarking emphasizes factual recall, offering limited insight into model performance on frontier and multimodal tasks. We assembled ***BioEVAL*** (BioEngineering Validation of AI and LLMs), a global, multi-institutional initiative designed to assess experimental reasoning capability across bioengineering (BE) subfields. ***BioEVAL*** spans 11 major BE subfields plus a set of uncategorized items, bringing together 22 research groups to create a PhD-level benchmark comprising 608 evaluation items: 1) 380 multiple-choice questions (MCQs, 359 retained after audit), 2) 218 literature synthesis tasks, and 3) 10 multimodal problems with experimental image interpretation. Benchmark items underwent authoring-group expert review and centralized quality control before evaluation. Following evaluation, a blinded cross-group consensus audit of the highest- and lowest-accuracy MCQ items flagged 21 questions for revision or removal; these were withheld, and all reported MCQ results are computed on the 359 retained items. We evaluated diverse cloud-scale foundation/multimodal models (*e.g.*, ChatGPT, Gemini, and Grok) and locally deployable models suitable for inference on consumer-grade GPUs. Models achieved the highest accuracy of up to 90% on MCQs, similarity score of 0.72 on literature synthesis, and accuracy of 80% on a small sample of multimodal reasoning questions, with substantial performance variation across subfields. Leaderboard rankings characterize current capabilities, limitations, and development priorities across the evaluated BE task categories. ***BioEVAL*** is maintained as an extensible benchmark with standardized protocols for continuing expert item contribution and model evaluation.

## II. Introduction

AI foundation models are rapidly becoming practical tools in bioengineering (BE), from protein structure prediction[1,2] and design[3,4] to literature synthesis[5,6] and experimental reasoning[7-9]. Related advances in generalist medical AI, multimodal biomedical AI, and AI-native scientific discovery further demonstrate the growing potential of foundation models across biomedical research and practices[10-12]. However, progress towards reliable real-world usage of these models relies on rigorous evaluation: without comprehensive domain-grounded benchmarks, we cannot fairly compare models, quantify readiness, diagnose failure modes, or direct the next generation of AI tool development[13,14]. Benchmarking is therefore a critical infrastructure, accelerating advancement across both general-purpose[15,16], and domain-specific models[17,18].

Despite growing interest in evaluating foundation and multimodal models for scientific use, most existing benchmarks remain dominated by exam-style, text-only tasks that emphasize factual recall or narrow question answering[19,20], although recent efforts such as LAB-Bench[21] and BioProBench[22] have begun to evaluate more research-oriented biological capabilities, including practical biology research tasks and biological protocol understanding and reasoning, as summarized in **Table 1** and **Table S1**. Datasets such as GPQA[23] and MedQA[24] measure performance on isolated multiple-choice or short-form reasoning prompts, providing useful baselines for general scientific knowledge and deductive reasoning. However, these conventional benchmark settings do not fully reflect the work of practicing bioengineers. Many remain only weakly research-grounded (*e.g.*, PhD-level, scenario-based items written by domain experts)[25]; they do not provide subfield-resolved insights into strengths and failure modes; and they largely omit the multimodal experimental evidence, such as plots, microscopy images, schematics, and assay outputs, that drives real experimental decisions[26,27]. ***BioEVAL*** complements these recent research-oriented benchmarks through its organization across bioengineering subfields, multi-institutional expert contributor network, literature-synthesis component, and pilot evaluation of experimental visual data.

**Table 1: Representative benchmark divisions across biology and life sciences, highlighting the lack of bioengineering-focused evaluation.**

| Benchmark Division | Representative Benchmarks | Primary Focus | Key Limitations for Bioengineering |
|---|---|---|---|
| Biomedical QA & Text | BioASQ[28], PubMedQA[29] | Textual knowledge, factual QA | Limited experimental reasoning; text-only |
| Biology Research & Procedural Reasoning | LAB-Bench[21], BioProBench[22] | Practical biology research capabilities; protocol understanding and procedural reasoning | Biology/protocol focused; limited bioengineering subfield organization and no integrated long-form literature-synthesis benchmark across BE domains |
| Clinical Reasoning | MedMCQA[30], MedQA[24], MultiMedQA[5] | Medical decision-making | Clinical focus, not experimental science |
| Chemistry & Molecular | ChemBench[31] | Molecular properties, reactions | Narrow molecular scope |
| General Science | GPQA[23], SciBench[32] | Broad graduate-level reasoning | Sparse bioengineering coverage |
| Medical Vision | GMAI-MMBench[33], OmniMedVQA[34], VQA-RAD[35], PathVQA[36], SLAKE[37] | Medical image understanding and visual question answering | Clinical/diagnostic focus; limited experimental and engineering reasoning |

**Table S1: Benchmark divisions across biology, life sciences, and engineering-oriented domains (with representative citations).**

| Benchmark Division | Representative Benchmarks | Primary Domain | Modalities | Typical Task Focus | Coverage of Bioengineering Experimental Reasoning |
|---|---|---|---|---|---|
| Biomedical Text & QA | PubMedQA[29] | Biology, Medicine | Text | Factual QA, literature comprehension | None |
| Biology Research & Procedural Reasoning | LAB-Bench[21], BioProBench[22] | Biology / Life Sciences | Text; figures/tables in LAB-Bench | Literature reasoning, protocol troubleshooting/generation, sequence manipulation, procedural reasoning | Relevant experimental/procedural reasoning, but not organized around broad BE subfields or integrated with BioEVAL-style literature synthesis and experimental-image evaluation |
| Clinical Reasoning | MedMCQA[30], MedQA[24], MultiMedQA[5] | Medicine | Text | Diagnosis, treatment decisions | None |
| Protein Structure & Function | ProteinGym[38] | Structural / protein biology | Sequence, structure | Structure and protein-fitness prediction | Indirect; little experimental design reasoning |
| Chemistry & Molecular Science | ChemBench[31] | Chemistry | Graphs, text | Property prediction, synthesis | None |
| Systems & Computational Biology | DREAM[39] | Systems biology | Tabular, simulation | Network inference | Limited (model-centric) |
| General Science Reasoning | MMLU[40], GPQA[23], SciBench[32] | Multi-domain science | Text | Graduate-level QA | Sparse and non-specific |
| Multimodal Science | MMMU[41] | Broad STEM | Image, text | Visual QA | Shallow bioengineering coverage |

To address these gaps, we introduce ***BioEVAL*** (BioEngineering Validation of AI and LLMs), a global, multi-institutional benchmarking initiative designed to evaluate large language and multimodal models on PhD-level BE tasks. Rather than measuring factual recall alone, ***BioEVAL*** jointly assesses domain knowledge, experimental reasoning, literature synthesis, multimodal interpretation, and semantic alignment between model-generated and expert-authored explanations. This enables a systematic evaluation of cloud-scale foundation and multimodal models (*e.g.*, ChatGPT, Gemini, and Grok), alongside locally deployable open-weight models, enabling comparison between cloud-accessed foundation models and models that can be run on consumer-grade GPUs, allowing direct comparison of capability and deployability across model scales using complementary metrics of accuracy and expert–model explanation-text similarity. ***BioEVAL*** provides task- and subfield-resolved leaderboards and diagnostic analyses that identify where current models perform consistently, where failure modes persist, and which BE capabilities require further validation before deployment in research settings. As a community resource, ***BioEVAL*** is intended to support reproducible model comparison, controlled benchmark expansion, and targeted dataset and model development toward more reliable AI-assisted experimental BE workflows.

## III. Results

### *III-1. BioEVAL* Benchmark Design and Evaluation Framework

The ***BioEVAL*** initiative assembled 22 research groups across North America, Europe, and Asia to construct a research-oriented benchmark spanning major bioengineering subfields (see prevalence of subfields in **Fig. S1**; Of the >21 BE subfields identified in the institutional survey, ***BioEVAL*** evaluated subfields for which at least two independent research groups participated). The resulting dataset comprised 608 expert-curated evaluation items across 11 predefined BE subfields, together with additional submissions outside these categories (**Fig. 1A**; **Methods V-2**).

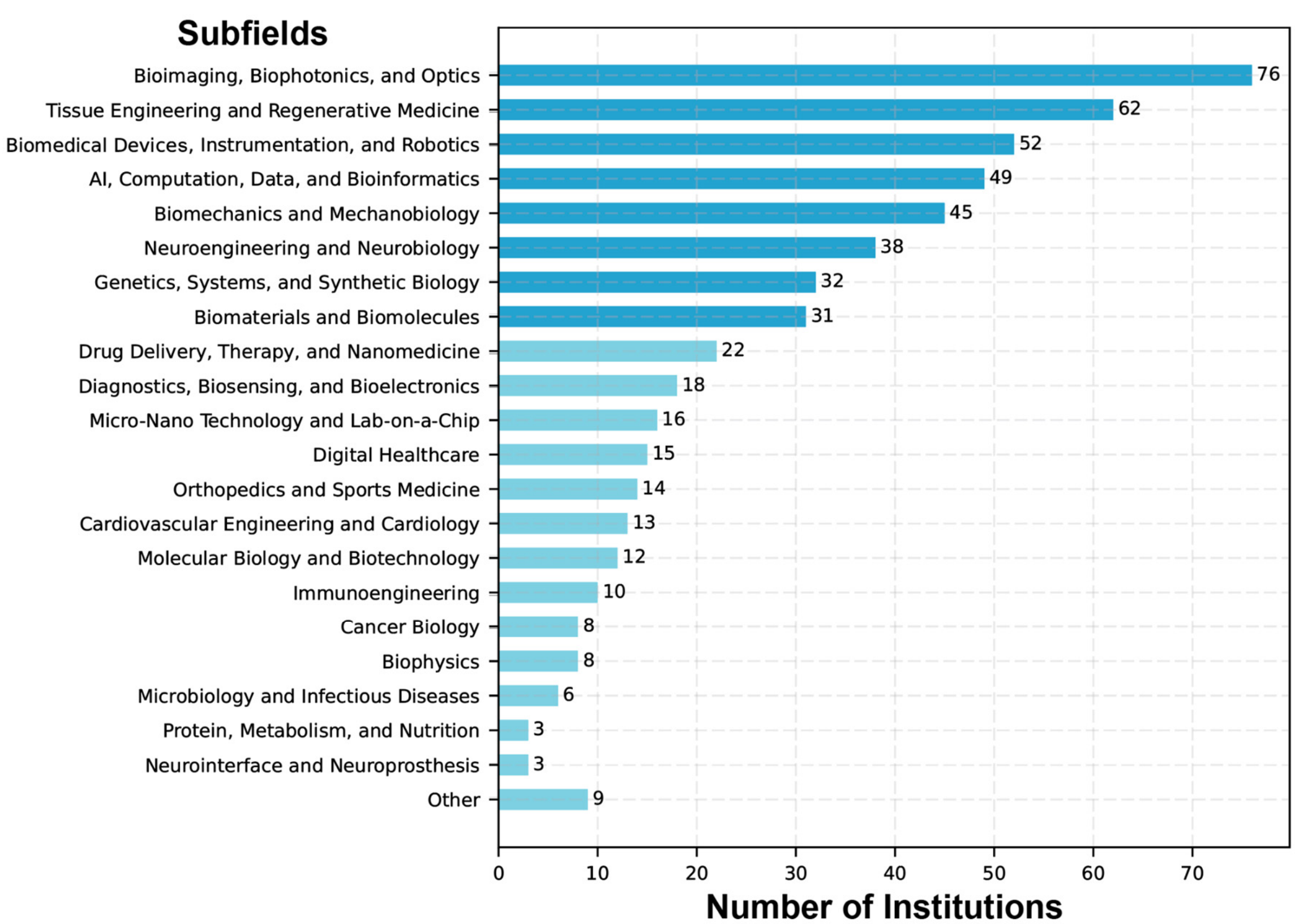

**Fig. S1** ***Institutional prevalence of BE subfields across globally ranked research institutions.*** Approximately 120 universities were selected from two widely recognized global university rankings (U.S. News and QS), and their BE departmental and faculty websites were manually reviewed to identify represented research subfields. The number of institutions in which each of more than 21 BE subfields was represented is shown. Dark cyan bars denote highly

represented subfields, while light cyan bars denote less frequently represented areas (The threshold is determined by the noticeable gap between the subfields *Biomaterials and Biomolecules* and *Drug Delivery, Therapy, and Nanomedicine*). The resulting distribution informed the prevalence of BE subfield research and, potentially, the availability of LLM training material.

These items are categorized into three complementary dimensions as follows (**Fig. 1A**).

1) Multiple-choice questions (**MCQs**), grounded in real BE research scenarios and experimental practices, that assess accuracy and semantic similarity between model-generated and expert-authored explanations.
2) Literature synthesis tasks (**L-Syn**) that evaluate long-form scientific article comprehension by prompting models to write publication-ready abstracts from abstract-withheld research papers.
3) Multimodal reasoning questions (**MRQs**) that assess experimental understanding grounded in visual data, including experimental images (*e.g.*, cell culture images), quantitative plots, and graphical illustrations.

Submitted items underwent expert review and centralized quality control to improve clarity, consistency, and answer-key integrity (**Fig. 1B**). Following model evaluation, a subset of MCQs selected by extreme model-response patterns underwent blinded cross-group expert audit, and items flagged for revision or removal were withheld from the reported results. The resulting benchmark therefore couples expert-authored content with task-specific evaluation of answer accuracy, expert-model explanation-text similarity, literature synthesis, and multimodal reasoning.

The ***BioEVAL*** framework further enables model performance analysis across multiple dimensions, including overall benchmark accuracy, cloud-scale versus edge-deployable model comparisons, task-specific leaderboards, and subfield-resolved performance heatmaps (**Fig. 1C**). These analyses provide a structured foundation for identifying the strengths and limitations of current AI systems, evaluating readiness for deployment in BE research workflows, and informing future development of AI-assisted scientific discovery and laboratory automation platforms.

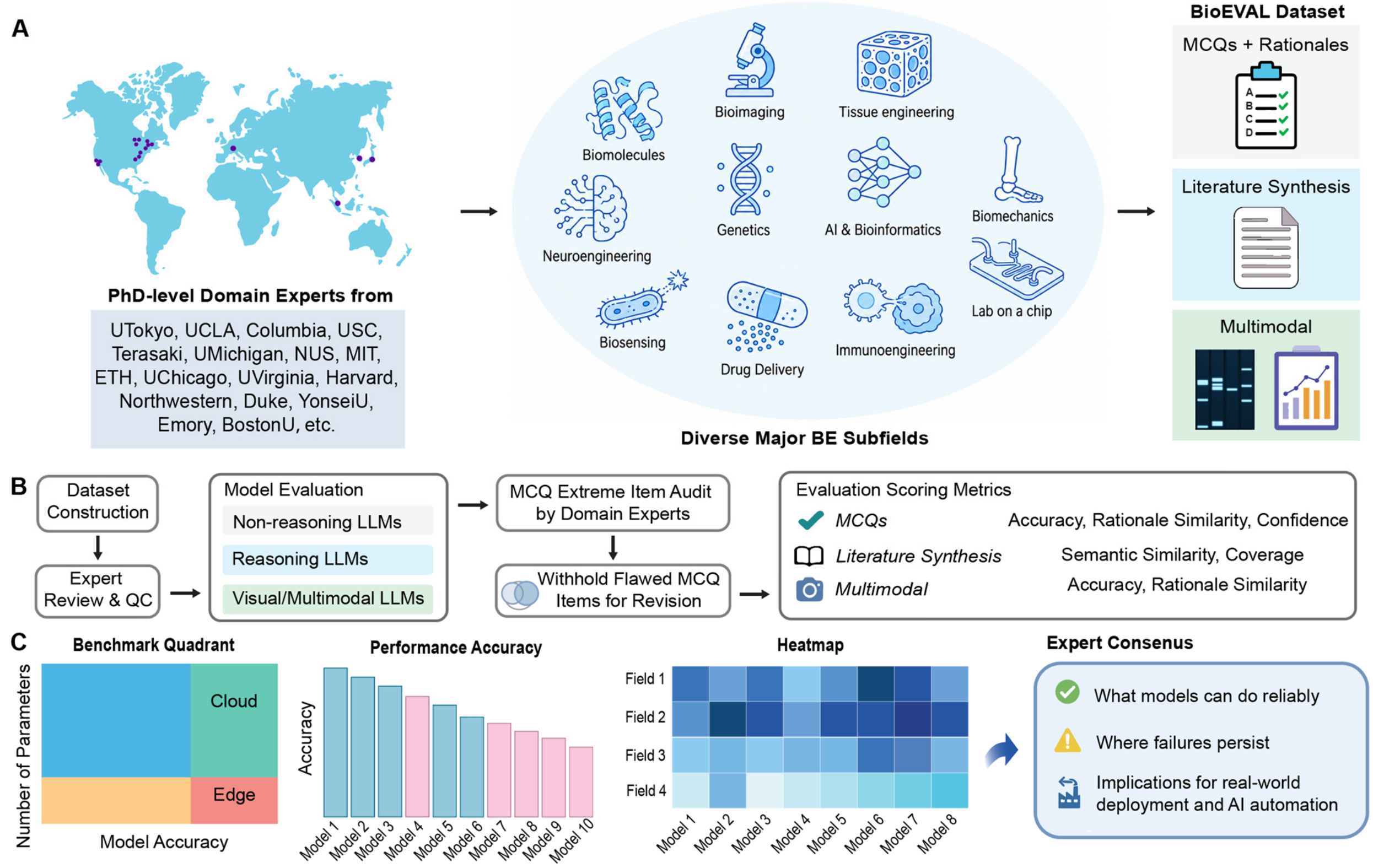

**Fig. 1** ***Overview of the BioEVAL benchmarking framework.*** (**A**) ***BioEVAL*** is a global, multi-institutional benchmarking initiative for evaluating large language models (LLMs) across diverse BE subfields. Research groups collaboratively curated 608 PhD-level evaluation items across major BE subfields. The benchmark comprises three evaluation categories: multiple-choice questions (MCQs) with expert rationales, literature synthesis from abstract-withheld articles, and multimodal reasoning problems that involve interpreting experimental visual data. (**B**) ***BioEVAL*** benchmark construction and evaluation workflow. Benchmark items were generated by subfield experts and underwent group-level review and centralized quality control before evaluation across non-reasoning LLMs, reasoning LLMs, and visual/multimodal models. Following evaluation, the highest- and lowest-accuracy MCQ items were pooled into a stratum-blinded 40-item audit set and reviewed by cross-group domain experts; the 21 items flagged for revision or removal were withheld, and reported MCQ results were recomputed on the 359 retained items. Task-specific metrics include accuracy and explanation-text similarity for MCQs; semantic similarity and coverage for literature synthesis; and accuracy and explanation-text similarity for multimodal reasoning tasks. (**C**) Model outputs are aggregated into comparative performance rankings based on accuracy and deployability, task- and subfield-resolved leaderboards, and heatmap-based visualizations across BE subfields. These analyses characterize model readiness, identify persistent limitations, and inform priorities for future AI-assisted BE research and laboratory automation.

## *III-2.* Evaluation of Subfield Knowledge and Experimental Reasoning Across Bioengineering MCQs

We first evaluated LLMs' ability to answer PhD-level multiple-choice questions (MCQs) across diverse BE subfields. An example of the evaluation pipeline, from prompting the LLM to answer an individual question, generating rationales, to evaluating the accuracy and similarity, is provided in **Fig. S2**. Across the full MCQ benchmark, frontier cloud-scale models achieved strong overall performance, with Gemini-2.5-Pro reaching the highest accuracy of the evaluated models at 90%, followed closely by several reasoning and non-reasoning frontier models with accuracies above 85% (**Fig. 2A**). These results suggest that current widely accessible LLMs have acquired substantial BE domain knowledge and can correctly answer many expert-curated questions involving experimental context and scientific reasoning.

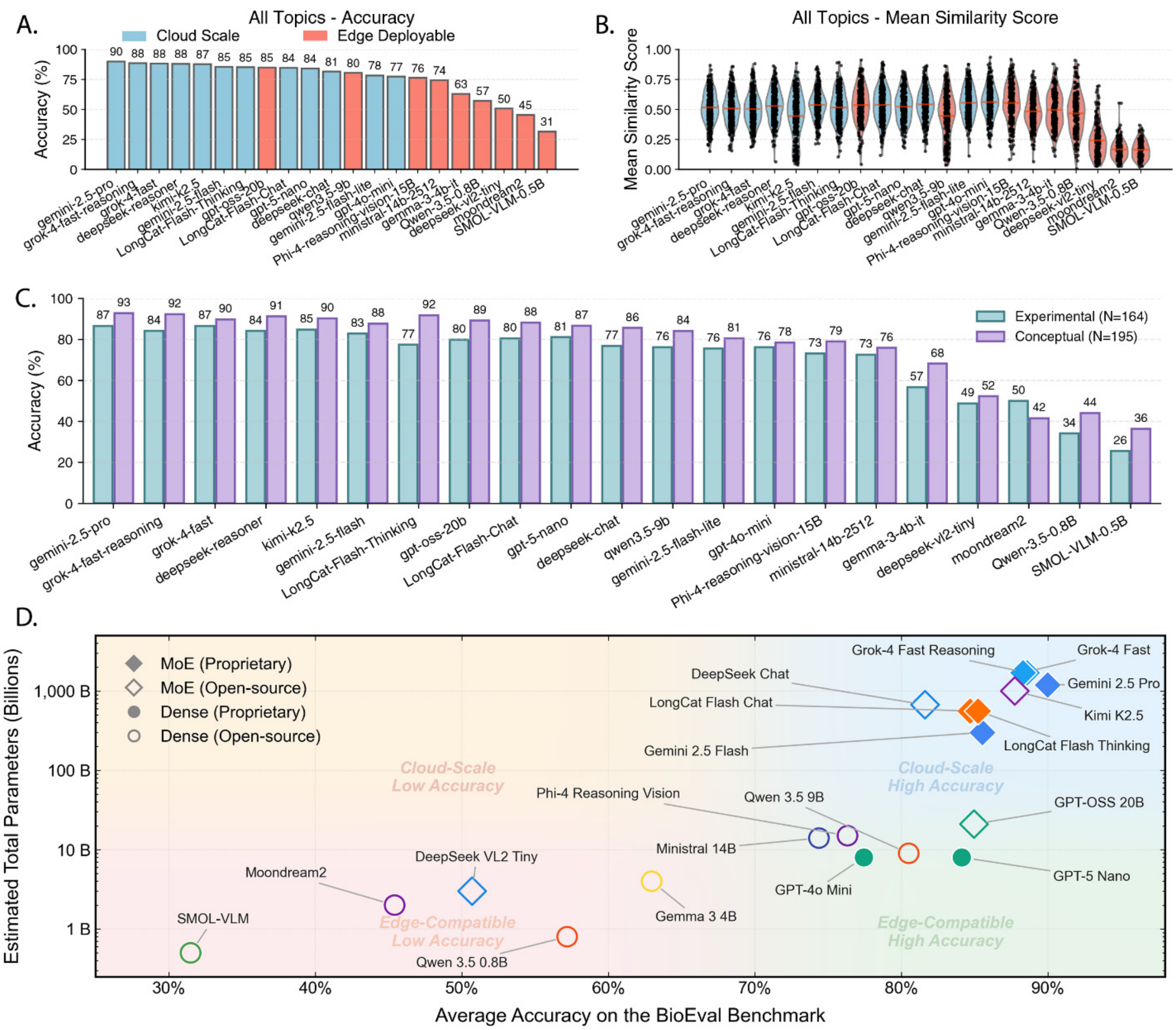

**Fig. 2** ***Evaluation of BE subfield knowledge and experimental reasoning across BioEVAL MCQs.*** (**A**) Overall accuracy of cloud-scale and edge-deployable LLMs across all ***BioEVAL*** multiple-choice questions (MCQs). Blue bars indicate cloud-scale models, and pink bars indicate edge-deployable models. (**B**) Distribution of expert-model explanation-text similarity scores. Domain experts provided explanations for why each answer option was correct or incorrect, and models were prompted, after answer selection, to generate structured explanations for all four options. Each model-generated explanation was compared with the corresponding expert-authored explanation, and the four option-level scores were averaged to obtain the per-question similarity score. This metric quantifies semantic agreement between explanation texts rather than the latent reasoning process that produced the model's answer. (**C**) Grouped bar analysis comparing model accuracy on classified experimental and conceptual MCQs (N = 164/195). Experimental questions consistently showed lower accuracy than conceptual questions across most tested models. (**D**) Relationship between model accuracy and total parameter count (estimated if not open-source). Models are organized by deployability and performance, with cloud-scale and edge-compatible regions indicated by color gradients. Marker shape indicates model architecture and accessibility, including mixture-of-experts (MoE) versus dense models and proprietary versus open-source models. In addition, rankings in this figure reflect observed benchmark accuracy and are descriptive; uncertainty intervals and formal pairwise significance testing were not performed in the current analysis.

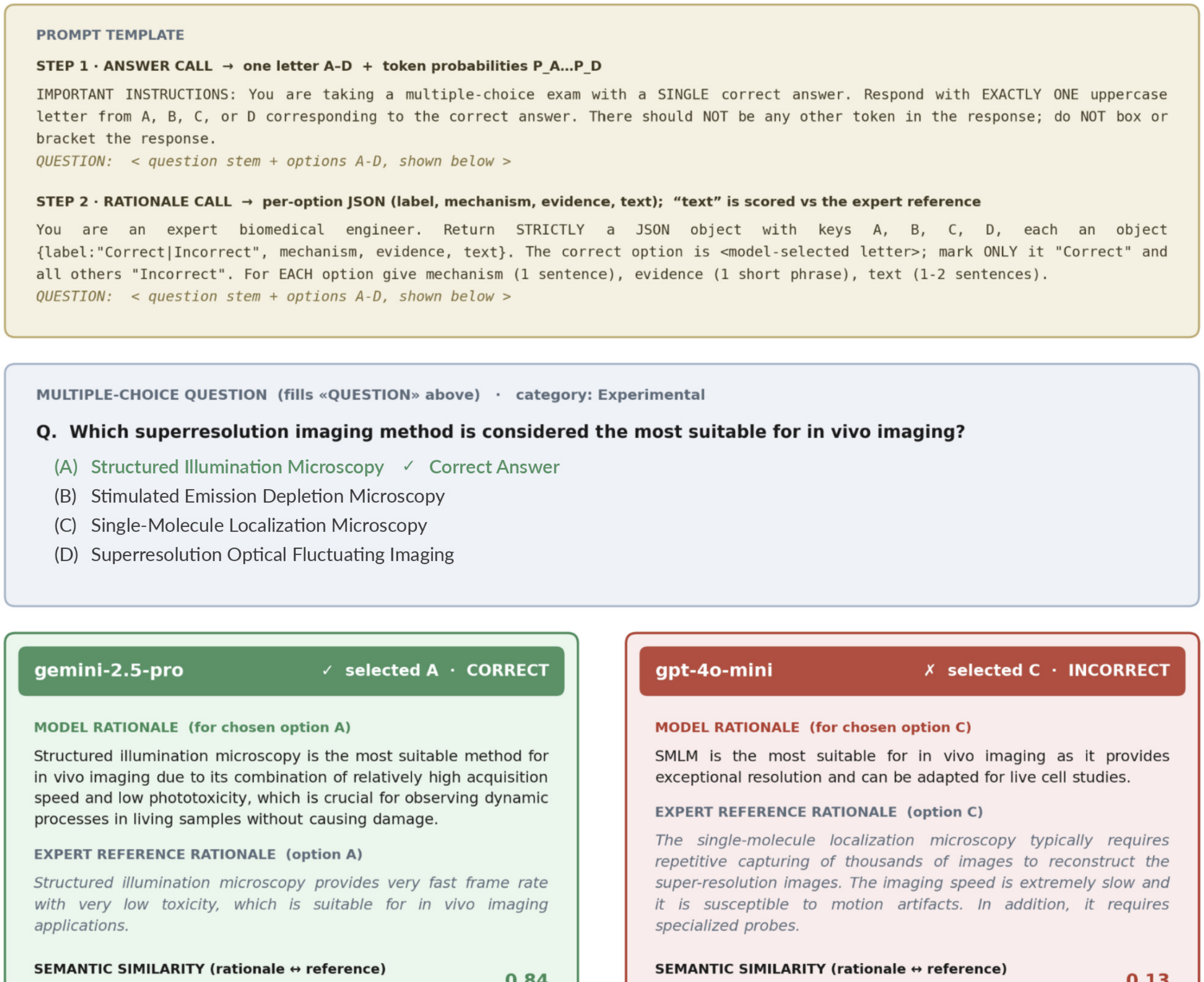

**Fig. S2** ***Domain knowledge evaluation pipeline with representative examples.*** Each multiple-choice question (MCQ) is presented to every model in two calls: an answer call returning a single letter (A–D) with token probabilities when supported, followed by a separate explanation call returning structured per-option justifications in JSON format (label, mechanism, evidence, and text). The explanation generated for each of the four answer options is compared with the corresponding expert-authored explanation using semantic similarity, and the four option-level scores are averaged to obtain a per-question expert–model explanation-text similarity score. The explanation call is performed after answer selection and therefore represents a post-hoc explanation rather than the latent reasoning process used to produce the answer. A representative question illustrates model answer selection and explanation-text similarity through this workflow. Aggregating answer correctness and explanation-text similarity across all 21 models and 359 questions produces the data summarized in **Fig. 2**.

Edge-deployable models showed a broader performance range but included several notable high-performing systems. GPT-oss-20B achieved 85% accuracy, approaching the performance of several larger cloud-scale models, while Qwen-3.5-9B reached 80% accuracy despite its compact model size (**Fig. 2A**). These findings indicate that local models can retain meaningful BE reasoning capability at substantially reduced scale. Such models may be particularly useful for privacy-sensitive research environments where experimental data or proprietary protocols may need to remain local.

In addition to answer accuracy, we evaluated the semantic alignment between model-generated and expert-authored explanations. During ***BioEVAL*** dataset construction, domain experts provided explanations describing why each answer option was correct or incorrect. After selecting an answer, models were separately prompted to generate structured explanations for all four options (**Fig. S2**). Each model-generated explanation was compared with the corresponding expert-authored explanation using semantic similarity metrics, and the four option-level scores were averaged to obtain a per-question expert-model explanation-text similarity score. Across models, explanation-text similarity distributions broadly followed the same trend as accuracy but showed substantial question-level variability, with the most accurate models generally centered near a similarity score of 0.5 (**Fig. 2B**). Several models exhibited broader distributions extending toward lower similarity values, indicating greater variability in semantic agreement with expert-authored explanations. For instance, the Qwen 0.8B and 9B models showed relatively high variance in explanation-text similarity. Despite comparatively high answer accuracy, these models showed lower mean explanation-text similarity, indicating that strong answer-selection performance did not necessarily coincide with similarly high semantic alignment of their post-hoc explanations with expert-authored explanations.

To further distinguish conceptual knowledge from experimental reasoning, MCQs were categorized (**Fig. S3**). Across most models, experimental questions yielded lower accuracy than conceptual questions except Moondream2 (**Fig. 2C**), indicating that models are generally more reliable when answering knowledge recall or principle understanding questions than when interpreting experimental scenarios. This performance gap suggests that experimental reasoning can be an area of further training and refinement for LLM deployment in practical, automated BE workflows, where models must reason across data sets and real-world uncertainties using biological, chemical, physical, and engineering frameworks rather than simply retrieve established facts.

| EXPERIMENTAL | CONCEPTUAL |
|---|---|
| **Q. How would you measure the internal pressure of a tumor during tumor growth?**<br>A. Laser ablation and retraction<br>B. AFM<br>**C. FRET based method ✓**<br>D. Micropipette aspiration | **Q. A node in a random network follows what distribution?**<br>**A. Poisson distribution ✓**<br>B. Power law distribution<br>C. Normal distribution<br>D. Bernoulli distribution |
| **Q. Which method below cannot generate iPSCs?**<br>A. Deliver transcription factors<br>B. Chemical induction<br>**C. Culture cells on top of matrigel ✓**<br>D. Direct delivery of reprogramming proteins | **Q. What is the main risk of using iPSCs in vivo?**<br>A. Hypoxia<br>B. Autoimmunity<br>**C. Teratoma formation ✓**<br>D. Infection |

**Fig. S3** ***Representative examples of experimental and conceptual MCQs.*** Experimental questions require practical or literature-informed reasoning about experimental design, execution, analysis, interpretation, or troubleshooting; conceptual questions primarily test established principles, properties, mechanisms, functions, or definitions. Two representative questions are shown per category with the correct option marked.

Further, we analyzed performance in relation to model scale and deployability (**Fig. 2D**). Frontier cloud-scale models occupied the high-accuracy region of the benchmark, consistent with their larger parameter counts, broader training data, and stronger reasoning capabilities. However, the performance of GPT-oss-20B and Qwen-3.5-9B demonstrates that locally deployable systems can achieve competitive observed accuracy while supporting inference on user-controlled hardware. Such deployment characteristics may be advantageous for privacy-sensitive research settings and motivate future evaluation in laboratory automation workflows. Together, these results indicate that tested LLMs already possess substantial BE domain knowledge with the remaining headroom concentrated in experimental-reasoning items and specific subfields.

Subfield-resolved analysis revealed substantial heterogeneity in model performance across BE research areas (**Fig. 3A**). *Diagnostics, Biosensing, and Bioelectronics*, *Neuroengineering and Neurobiology*, and *Immunoengineering* questions consistently yielded high accuracy across many models, with several models reaching or approaching perfect performance. In contrast, MCQs in *Biomaterials and Biomolecules*, *Genetics, Systems and Synthetic Biology*, and *Drug Delivery, Therapy, and Nanomedicine* showed markedly lower performance, with many models falling below 70% accuracy and some compact models performing below 50%. To evaluate whether this heterogeneity could be partially explained by question type, we further examined the fraction of experimental versus conceptual questions within each subfield (**Fig. S4**). In the current dataset, subfields with a higher proportion of experimental questions tended to show lower observed accuracy, consistent with the descriptive experimental-versus-conceptual comparison reported above. However, this relationship was not absolute. For example, *Micro-Nano Technology and*

*Lab-on-a-Chip* and *Diagnostics, Biosensing, and Bioelectronics* contained high fractions of experimental questions but still achieved strong model performance, whereas *Genetics, Systems and Synthetic Biology* contained a high proportion of conceptual questions but showed relatively low accuracy. These exceptions suggest that subfield performance is not explained by experimental/conceptual composition alone and may also reflect differences in question difficulty, contributor-specific question design, and model training-data representation.

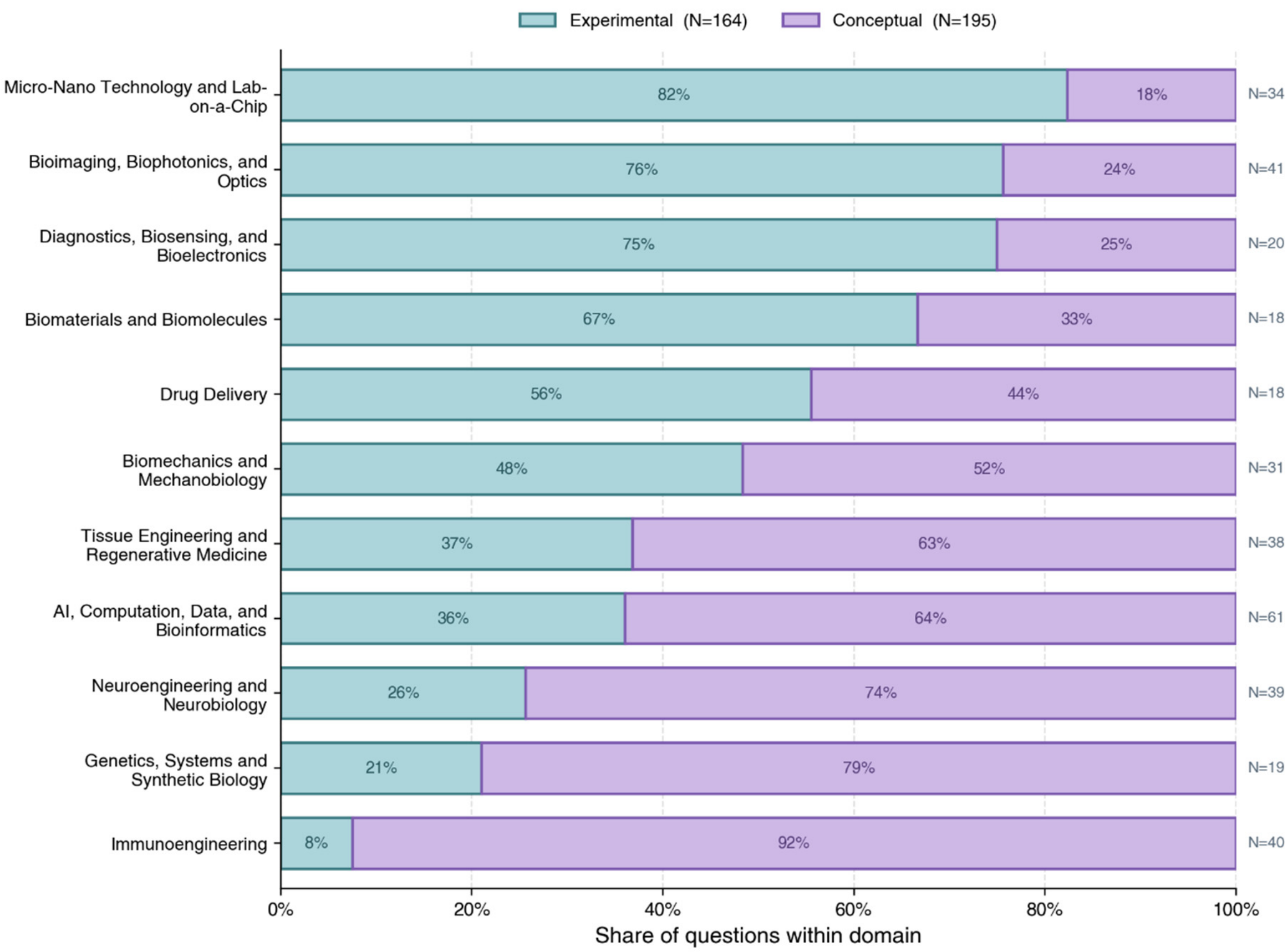

**Fig. S4** ***Distribution of experimental and conceptual question types within each bioengineering domain.***

To determine whether these subfield-level trends were driven by broad topic effects or by specific difficult questions, we analyzed per-question correctness patterns. Each model was represented by a binary correctness vector across the 359 MCQs, and questions were grouped by subfield and defined as the fraction of models that answered each question correctly (**Fig. 3B**). Model errors were not randomly distributed across the benchmark.

To investigate whether extreme model-response patterns could identify benchmark items warranting additional review, we performed a focused analysis of the 20 MCQs with the lowest model accuracy (**Fig. S5**). These items were concentrated in several subfields, including *Bioimaging, Biophotonics, and Optics* and *AI, Computation, Data, and Bioinformatics* (**Fig. S5A**). Because the subsequent post-evaluation cross-group audit identified potential problems in many of these items, their subfield distribution should not be interpreted as evidence that models are intrinsically weaker in these domains. Instead, this analysis illustrates how unusually low cross-model accuracy can serve as a diagnostic signal for benchmark quality control.

The hardest items also differed descriptively in their framing. Consistent with (**Fig. 2C**), experimental questions were enriched among the 20 lowest-accuracy items compared with the full MCQ set (**Fig. S5B**), whereas superlative wording and negation or exception-style classification showed smaller differences. Given that many of these items were subsequently flagged for revision, these patterns are best interpreted as characteristics associated with items requiring additional review rather than as established causes of model failure.

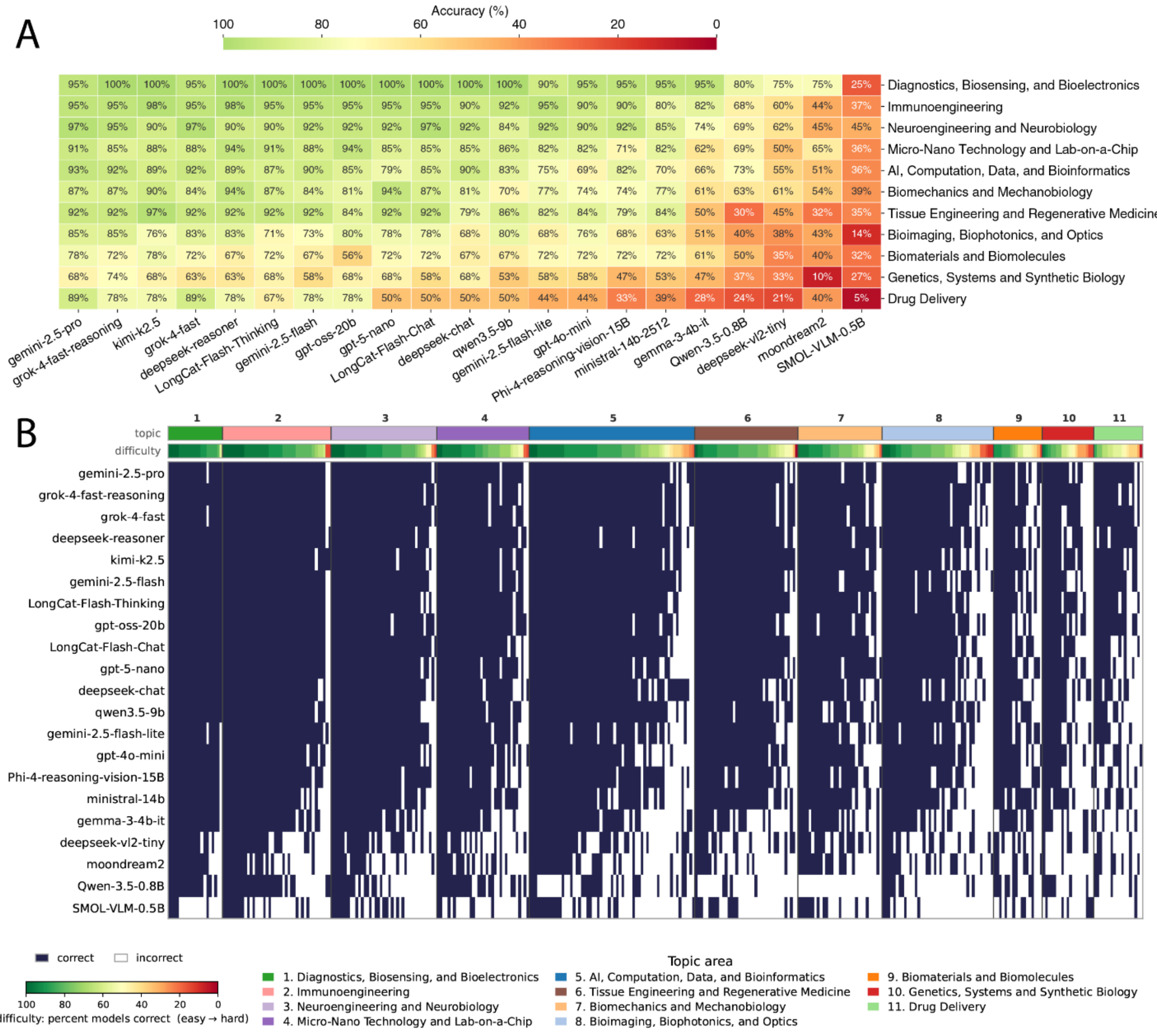

**Fig. 3** ***Subfield and Per-question accuracy of the tested models on the BioEVAL dataset.*** (**A**) Subfield-resolved heatmap of MCQ accuracy across BE subfields and evaluated models excluding uncategorized questions. (**B**) Each cell shows whether a model answered a question correctly (dark) or incorrectly (white). Rows are models ordered by overall accuracy (most accurate at top); columns are questions grouped into 11 topic areas and ordered from easy to hard within each group. The top strips annotate each question's topic area and difficulty (fraction of models correct). Because the number of MCQs varies across subfields (**Fig. S4**), subfield-level accuracies are descriptive and are more sensitive to individual questions in domains with smaller item counts.

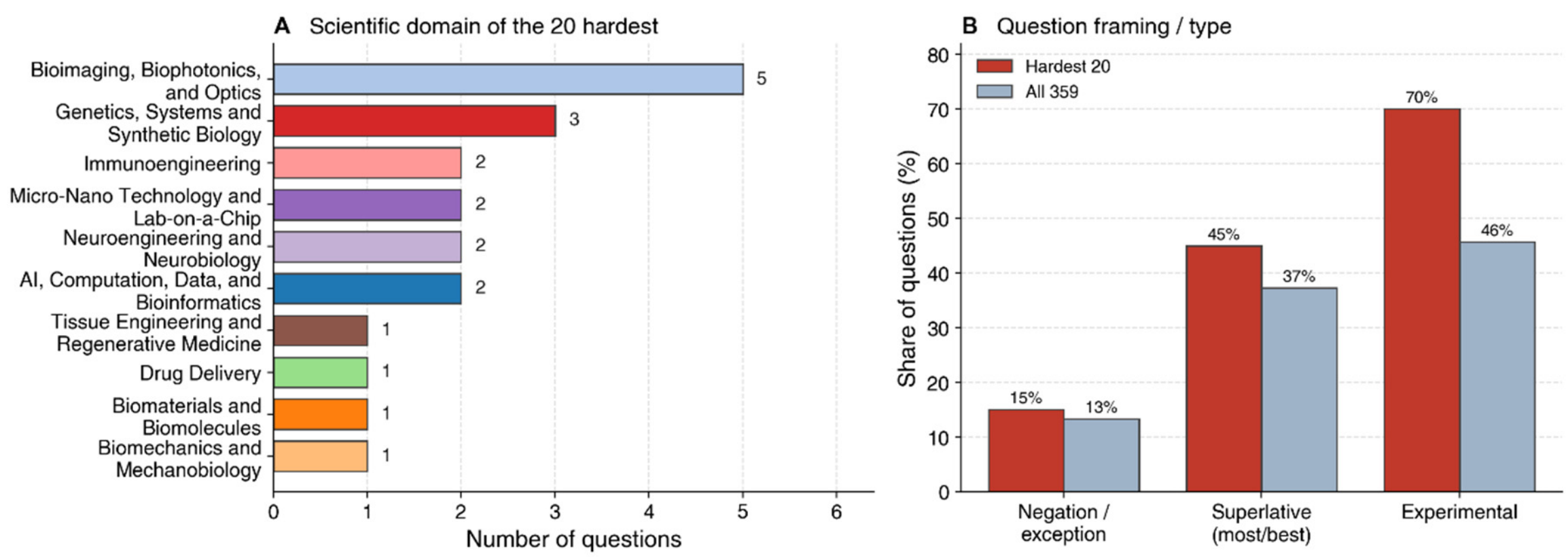

**Fig. S5** ***Model-response patterns among the 20 lowest-accuracy BioEVAL MCQs.*** (**A**) Distribution of 20 items across scientific domains. (**B**) Descriptive comparison of question framing and type between the lowest-accuracy 20 items and the full MCQ benchmark, including negation/exception phrasing, superlative phrasing, and experimental-question classification. Because subsequent expert reviews identified potential item-quality issues in many of these questions, these patterns are interpreted primarily as diagnostic signals for benchmark review rather than as evidence of shared model failure mechanisms.

Seventeen domain experts returned feedback on the 20 lowest accuracy items that were randomly mixed with the 20 highest accuracy items to avoid biasing responses (40 audited MCQ items in total). The domain experts submitted responses in seven distinct formats casting 231 votes in total (median 5 votes per item, range 2-12). The post-evaluation audit recommended retaining 19 items (47.5%) unchanged, revising 19 (47.5%), and removing 2 (5.0%) (**Fig. S6A**). The audit was conducted after model inference, and the models were not rerun. Because MCQ accuracy is computed per item, results were instead recomputed over the retained subset: the 19 items flagged for revision and the 2 flagged for removal were withheld from scoring, and **Figs. 2**, **3**, **6**, and **S5** report performance on the 359 retained questions. The 19 items returned for revision are held pending expert revision and are intended for reinstatement in a subsequent release.

Requests for change were strongly concentrated in the difficult stratum (**Fig. S6B**). Of the 20 items answered correctly by the fewest models, the audit recommended retaining 4, revising 14, and removing 2; among the 20 items answered correctly by all models, it recommended retaining 15, revising 5, and removing none. The difference in the proportion of items requiring any change was significant (16/20 versus 5/20; two-sided Fisher's exact test, $p = 0.0012$). Both removed items originated in the difficult stratum, although with only two removals this difference was not on its own statistically resolvable (2/20 versus 0/20, $p = 0.49$).

The same asymmetry was evident at the level of individual votes (**Fig. S6C**). Reviewers endorsed retention in 78% of the 122 votes cast on universally answered items, but in only 36% of the 109 votes cast on the hardest items. All seven unassigned "revise or remove" verdicts fell within the hardest stratum. In their free-text justifications, reviewers attributed a substantial proportion of near-universal model failures to ambiguous phrasing, unclear superlatives (for example "least important", "most effectively enhances"), or the presence of more than one defensible answer, rather than to the depth of domain knowledge required.

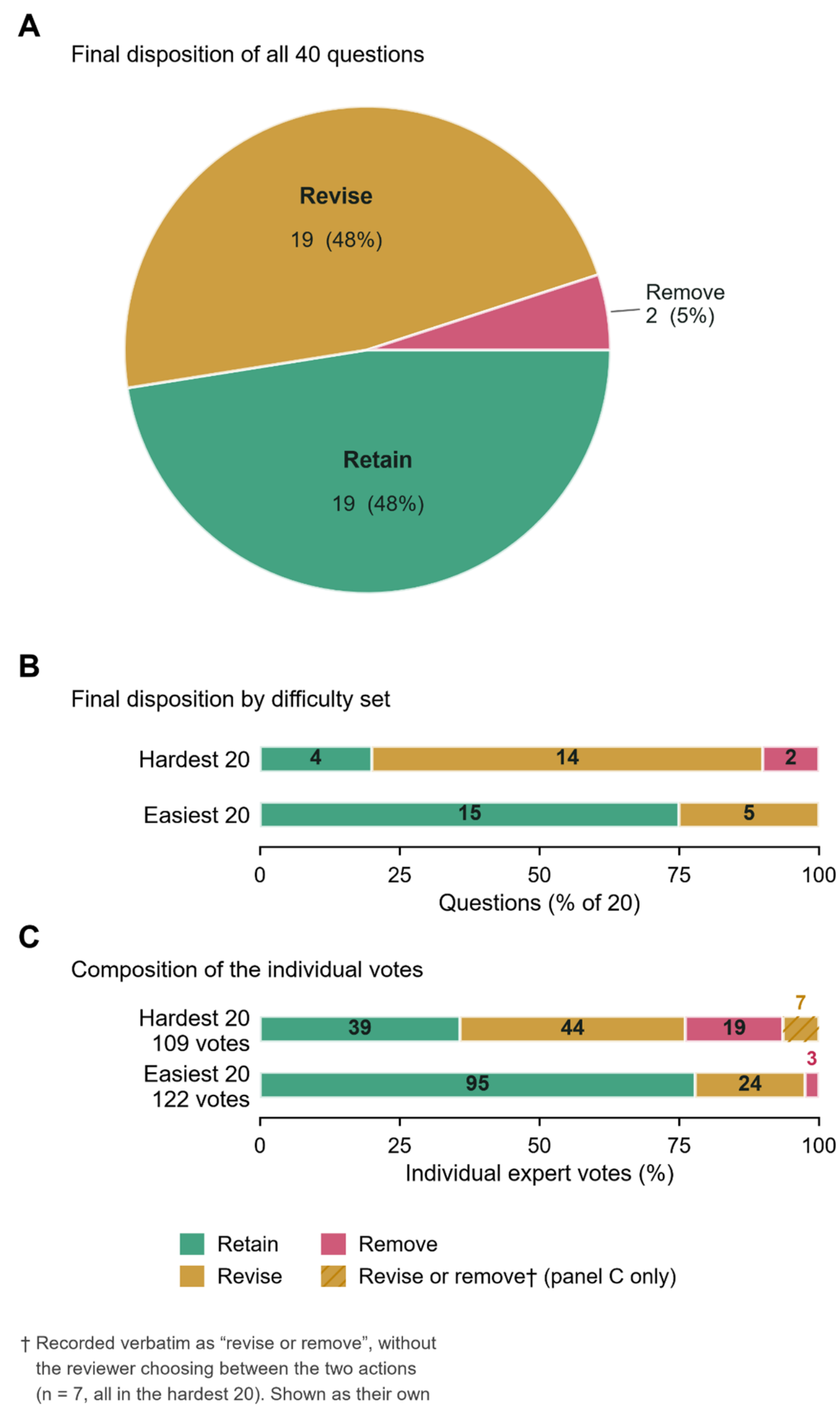

**Fig. S6** ***Post-evaluation cross-group audit of 40 selected MCQ items.*** Seventeen groups with domain experts reviewed a single shuffled list combining the 20 questions answered correctly by the fewest of 21 evaluated models (Hardest 20; 0-3/21 correct) with the 20 answered correctly by all models (Easiest 20; 21/21 correct), casting 231 votes in total (median 5 votes per question, range 2-12). (**A**) Audit recommendations across all 40 questions. (**B**) Audit recommendations within each difficulty set; questions requiring any change (revise or remove) are concentrated in the hardest 20 (16/20 vs 5/20; two-sided Fisher's exact test, $p = 0.0012$), whereas both removals, being only two, do not by themselves establish a difference ($p = 0.49$). (**C**) Composition of the individual votes within each difficulty set.

## *III-3.* Literature Synthesis Performance Across BE Subfields

To evaluate the ability of LLMs to comprehend and synthesize BE scientific literature, we developed a literature synthesis (L-Syn) benchmark comprising 218 abstract-withheld full-text primary research articles selected by domain experts around active research questions across BE subfields; models were tasked with generating publication-style abstracts (**Fig. 4A**). Generated abstracts were compared against the original author-written abstracts using embedding-based semantic similarity metrics that jointly quantify semantic precision and coverage. An example of evaluating the LLM on an individual abstract-withheld article is provided in **Fig. S7**.

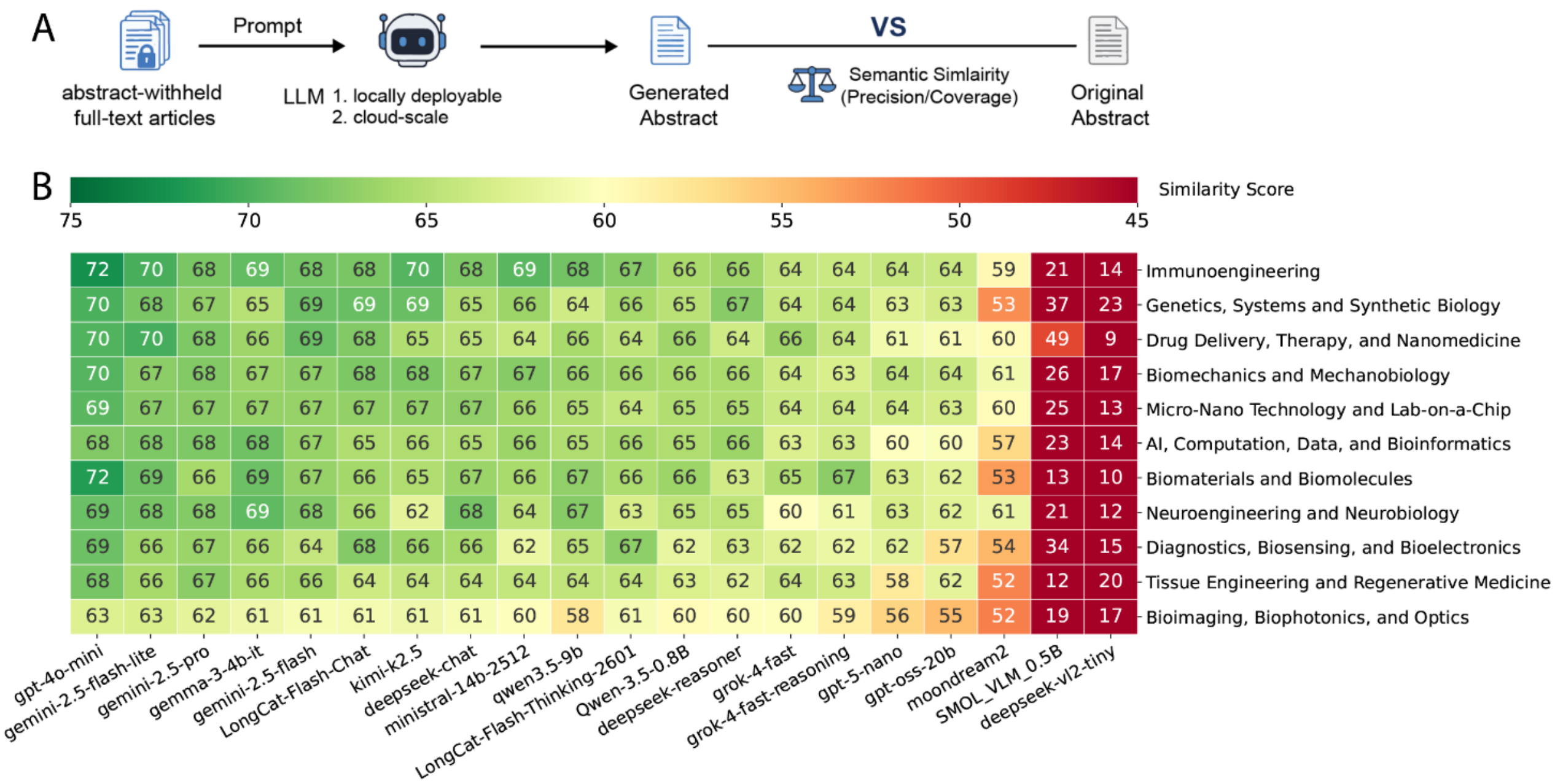

**Fig. 4** ***Evaluation of literature synthesis performance in BioEVAL.*** (**A**) Overview of the literature synthesis (L-Syn) benchmark. Abstract-withheld full-text articles (capped at 50,000 tokens as described in Methods V-7) are provided to LLMs prompted to generate abstracts. Generated abstracts were compared with the original author-written abstracts using embeddings to assess semantic precision and coverage. (**B**) Heatmap of average L-Syn performance across the 11 categorized bioengineering subfields for cloud-scale and edge-deployable LLMs. Similarity scores range from low (red) to high (green), with higher values indicating stronger agreement between generated and reference abstracts. Differences among BE subfields indicate inconsistent L-Syn capability across subfields.

PROMPT TEMPLATE

**ABSTRACT CALL → free-text abstract (150–250 words)**

You are a leading scientist with expertise in the field relevant to the provided manuscript. Based on the following full-text research article, generate a publication-ready abstract that is concise, comprehensive, and follows standard academic abstract structure. The abstract should: summarize the key research question, methods, results, and conclusions; be approximately 150-250 words; use clear, professional scientific language; highlight the main contributions and findings.

*Full-text article: < full text of the research article, abstract removed >*

**REFERENCE ABSTRACT · MicroRNA-21 preserves the fibrotic mechanical memory of mesenchymal stem cells (Nat. Mat., 2017)**

Expansion on stiff culture substrates activates pro-fibrotic cell programs that are retained by mechanical memory. Here, we show that priming on physiologically soft silicone substrates suppresses fibrogenesis and desensitizes mesenchymal stem cells (MSCs) against subsequent mechanical activation in vitro and in vivo, and identify the microRNA miR-21 as a long-term memory keeper of the fibrogenic program in MSCs. During stiff priming, miR-21 levels were gradually increased by continued regulation through the acutely mechanosensitive myocardin-related transcription factor-A (MRTF-A/MLK-1) and remained high over 2 weeks after removal of the mechanical stimulus. Knocking down miR-21 once by the end of the stiff-priming period was sufficient to erase the mechanical memory and sensitize MSCs to subsequent exposure to soft substrates. Soft priming and erasing mechanical memory following cell culture expansion protects MSCs from fibrogenesis in the host wound environment and increases the chances for success of MSC therapy in tissue-repair applications.

**HIGH similarity: LongCat-Flash-Thinking**

FINAL SIMILARITY SCORE 0.69

GENERATED ABSTRACT

Mechanical memory in mesenchymal stem cells (MSCs) influences therapeutic outcomes, yet its molecular mechanisms remain unclear. We investigated whether substrate stiffness during culture expansion imprints persistent fibrogenic memory and identified microRNA-21 (miR-21) as a key regulator. Rat bone marrow MSCs were primed on physiologically soft (5 kPa) or pathologically stiff (100 kPa) silicone substrates for up to five passages. Soft-priming suppressed α-smooth muscle actin (α-SMA) expression, enhanced clonogenicity, and preserved multipotency, while stiff-priming induced progressive myofibroblast activation that persisted for at least two passages after switching to soft substrates. miR-21 expression increased 2.5-fold on stiff substrates and remained stable after mechanical changes, whereas myocardin-related transcription factor-A (MRTF-A) acutely regulated miR-21 transcription but was reset after passaging. Chromatin immunoprecipitation confirmed SRF binding to the miR-21 promoter CArG box. Critically, a single miR-21 knockdown in stiff-primed MSCs erased mechanical memory, restoring responsiveness to soft substrates. In a splinted rat wound model, soft-primed MSCs reduced scar contracture by 2-fold and myofibroblast formation by 50% compared to stiff-primed MSCs, which exacerbated scarring. Memory-erased MSCs phenocopied soft-primed cells, enhancing angiogenesis and reducing fibrosis. These findings demonstrate that miR-21 serves as a long-term memory keeper for fibrogenic programs, and soft-priming or miR-21 manipulation can optimize MSC therapy for regenerative wound repair.

**LOW similarity: gemma-3-4b-it**

FINAL SIMILARITY SCORE 0.38

GENERATED ABSTRACT

Mechanical memory profoundly influences the behavior of mesenchymal stem cells (MSCs), impacting their fibrogenic potential and long-term responses to mechanical stimuli. This study demonstrates that MSCs cultured directly on physiologically soft silicone substrates acquire a robust mechanical memory, retaining enhanced fibrogenic characteristics and altered lineage differentiation potential even after transplantation and subsequent exposure to stiffer environments. Unlike MSCs expanded on conventional tissue culture plastic, soft-primed MSCs exhibit reduced α-SMA expression, increased colony formation, and a heightened propensity for myofibroblast differentiation. Furthermore, we identified microRNA-21 (miR-21) as a key regulator of this mechanical memory, with miR-21 levels gradually increasing during soft substrate priming and being stably maintained across passages, independent of substrate stiffness changes. MRTF-A, a mechanosensor, directly controls miR-21 expression, establishing a critical link between substrate mechanics and fibrotic programming. Importantly, this mechanical memory persists for at least two passages, suggesting a potential therapeutic target for mitigating fibrosis. Finally, transplantation of memory-erased, soft-primed MSCs into a wound model resulted in reduced scar formation and tissue contraction compared to MSCs primed on stiffer substrates, highlighting the clinical relevance of this mechanical memory phenomenon. These findings establish a novel mechanism for controlling MSC behavior and offer a promising strategy for enhancing the efficacy of MSC-based therapies for tissue repair.

**Fig. S7** ***A representative example of the L-Syn abstract-generation evaluation and semantic similarity.*** A model is given a paper's full text (abstract removed) and is prompted to write an abstract. The output is then scored against the real abstract using sentence-embedding similarity metrics. Underlined regions show either matched (left) or mismatched (right) understanding of the paper's results.

Across all evaluated models, literature synthesis performance exhibited substantial variability, with average similarity scores ranging from 0.09 to 0.72 across BE subfields (**Fig. 4B**). Among the tested models, GPT-4o-mini and Gemini-2.5-Flash-Lite demonstrated the strongest and most consistent performance, achieving similarity scores between 0.63 and 0.72 across all evaluated subfields. Several other cloud-scale models, including Gemini-2.5-Pro, Gemini-2.5-Flash, LongCat-Flash-Chat, Kimi-K2.5, and DeepSeek-Chat, also maintained relatively stable performance with average scores clustered ~0.61–0.70.

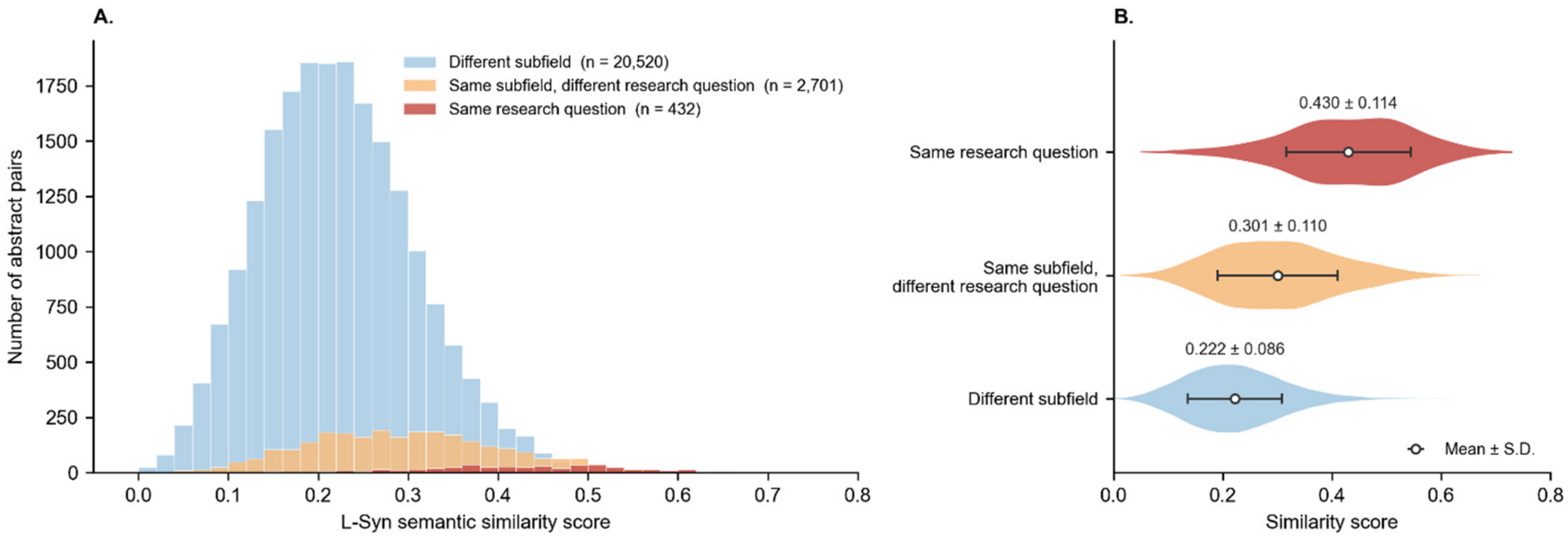

**Fig. S8** ***Reference distribution of the L-Syn semantic similarity metric across mismatched author-written abstracts.*** All mismatched pairs among the 218 reference abstracts were scored using the same semantic similarity procedure applied to model-generated abstracts and stratified by relatedness. (**A**) Histogram of individual abstract-pair similarity scores by relatedness tier. (**B**) Per-tier distributions with mean ± S.D. Different-subfield pairs scored 0.222 ± 0.086 (n = 20,710), same-subfield pairs from different research questions scored 0.301 ± 0.109 (n = 2,724), and non-matching articles associated with the same research question scored 0.430 ± 0.114 (n = 436). These distributions characterize the background similarity observed between distinct scientific articles and are not used as a matched null distribution for the averaged model-subfield scores in **Fig. 4B**.

To contextualize the scale of the L-Syn similarity metric at the individual-article level, we scored mismatched pairs of author-written abstracts (**Fig. S8**). Different-subfield pairs scored 0.222 ± 0.086, same-subfield pairs from different research questions scored 0.301 ± 0.109, and non-matching articles associated with the same research question scored 0.43 ± 0.114. An example of five articles nominated under a single research question, and the similarity between two of their abstracts, is shown in **Fig. S9.** These distributions provide a reference for the level of similarity that can arise between distinct scientific articles of varying topical relatedness.

**A. Research question**

*How do time-dependent mechanical properties of tissues (viscoelasticity, plasticity) affect long-term cell behaviors and adaptation?*

**Source papers answering this question in the L_Syn dataset:**

1. **Mechanical memory and dosing influence stem cell fate (Nat. Mat., 2014) (shown in B)**
2. The mechanical memory of lung myofibroblasts (Integr. Biol., 2012)
3. MicroRNA-21 preserves the fibrotic mechanical memory of mesenchymal stem cells (Nat. Mat., 2017)
4. **Linking cell mechanical memory and cancer metastasis (Nat. Rev. Cancer, 2024) (shown in B)**
5. Biophysical Regulation of Chromatin Architecture Instills a Mechanical Memory in Mesenchymal Stem Cells (Sci. Rep, 2015)

**B. Example abstract pair from this group — L-Syn similarity score 0.500**

**PAPER 1 · Mechanical memory and dosing influence stem cell fate (Nat. Mat., 2014)**

We investigated whether stem cells remember past physical signals and whether these can be exploited to dose cells mechanically. We found that the activation of the Yes-associated protein (YAP) and transcriptional coactivator with PDZ-binding domain (TAZ) as well as the pre-osteogenic transcription factor RUNX2 in human mesenchymal stem cells (hMSCs) cultured on soft poly(ethylene glycol) (PEG) hydrogels (Young's modulus E ~ 2 kPa) depended on previous culture time on stiff tissue culture polystyrene (TCPS; E ~ 3 GPa). In addition, mechanical dosing of hMSCs cultured on initially stiff (E ~ 10 kPa) and then soft (E ~ 2 kPa) phototunable PEG hydrogels resulted in either reversible or--above a threshold mechanical dose--irreversible activation of YAP/TAZ and RUNX2. We also found that increased mechanical dosing on supraphysiologically stiff TCPS biases hMSCs towards osteogenic differentiation. We conclude that stem cells possess mechanical memory--with YAP/TAZ acting as an intracellular mechanical rheostat--that stores information from past physical environments and influences the cells' fate.

**PAPER 4 · Linking cell mechanical memory and cancer metastasis (Nat. Rev. Cancer, 2024)**

Metastasis causes most cancer-related deaths; however, the efficacy of anti-metastatic drugs is limited by incomplete understanding of the biological mechanisms that drive metastasis. Focusing on the mechanics of metastasis, we propose that the ability of tumour cells to survive the metastatic process is enhanced by mechanical stresses in the primary tumour microenvironment that select for well-adapted cells. In this Perspective, we suggest that biophysical adaptations favourable for metastasis are retained via mechanical memory, such that the extent of memory is influenced by both the magnitude and duration of the mechanical stress. Among the mechanical cues present in the primary tumour microenvironment, we focus on high matrix stiffness to illustrate how it alters tumour cell proliferation, survival, secretion of molecular factors, force generation, deformability, migration and invasion. We particularly centre our discussion on potential mechanisms of mechanical memory formation and retention via mechanotransduction and persistent epigenetic changes. Indeed, we propose that the biophysical adaptations that are induced by this process are retained throughout the metastatic process to improve tumour cell extravasation, survival and colonization in the distant organ. Deciphering mechanical memory mechanisms will be key to discovering a new class of anti-metastatic drugs.

**Fig. S9** ***Representative example of articles nominated under a shared research question in the L-Syn dataset.*** (**A**) One research question contributed by a participating group, together with the five primary research papers nominated by domain experts as addressing the same scientific problem. Articles nominated under a common research question constitute the same-research-question tier of the mismatched-pair reference distribution in **Fig. S8** (n = 436 pairs, 0.43 ± 0.114). (**B**) Author-written abstracts of two articles from this group (papers 1 and 4), scored against each other using the same semantic similarity procedure applied to model-generated abstracts (**Methods V-6**). Despite addressing a shared theme of mechanical memory in cells, the pair scored 0.500, within the upper range of the same-research-question tier but well below the scores obtained by well-performing models against their matched reference abstracts (**Fig. 4B**), suggesting that scores in this range may reflect shared subject matter rather than recovery of a specific article's findings.

Performance differences were observed across BE subfields. *Immunoengineering* achieved the highest overall similarity scores, with top-performing models reaching 0.72, followed closely by *Genetics, Systems and Synthetic Biology*, and *Drug Delivery, Therapy, and Nanomedicine*. In contrast, *Bioimaging, Biophotonics, and Optics* generally yielded lower similarity scores, clustered ~0.52–0.63, suggesting that L-Syn in image-intensive and instrumentation-focused domains may present additional challenges. Notably, substantial performance variation was observed between models within the same subfield. Because the evaluated models differ simultaneously in architecture, training data, scale, and inference strategy, the present analysis does not identify which model characteristics account for these differences.

A separation was observed between frontier cloud-scale models and smaller edge-deployable models. While most cloud-scale systems consistently achieved similarity scores above 0.60, edge-deployable models such as GPT-oss-20B showed slight degradation across all subfields, producing similarity scores below 0.6, for example, 0.56 in *Bioimaging, Biophotonics, and Optics*. In addition, smaller models such as SMOL_VLM_0.5B and deepseek_vl2_tiny (1B parameters) consistently produced inferior similarity scores, with deepseek_vl2_tiny deteriorating to 0.09 for L-Syn in *Drug Delivery, Therapy, and Nanomedicine*. These results demonstrate substantial variation in L-Syn performance across evaluated models, but the present study does not isolate the effects of model scale, architecture, training data, or inference strategy on this variation.

## *III-4.* Multimodal Reasoning Performance on Experimental BE Data

We next evaluated LLMs' ability to interpret experimental visual data using a pilot ***BioEVAL*** multimodal reasoning benchmark. Unlike text-only MCQs, these questions require models to jointly analyze an accompanying image or graph and apply relevant BE domain knowledge during the reasoning. The current pilot set contains 10 expert-curated multimodal questions covering representative experimental data types, including qPCR amplification curves, SDS–PAGE gels, bacterial growth curves, fed-batch fermentation profiles, mycoplasma PCR quality-control gels, Sanger sequencing chromatograms, adherent mammalian cell culture images, and enzyme-kinetics plots. Some examples are provided in **Note S1**. Because of the limited number of items, this analysis is intended as an initial assessment of multimodal reasoning capability rather than used for a definitive ranking of model performance.

## Note S1 *Examples of Multimodal Questions (Images Excluded)*

**Q1 - qPCR Amplification Curves (ΔCt and Fold Difference)**
A qPCR plot is shown for several samples containing different concentrations of a nucleic acid target. Each line type and color refers to replicates of a specific condition. Assuming similar amplification efficiency and ideal doubling each cycle, what is the approximate fold difference in starting template quantity between the red and black groups?
A. 4-fold
B. 8-fold
C. 64-fold
D. 256-fold

**Correct Answer**
**C. 64-fold**

**Detailed Option Analysis**
**A. 4-fold – Incorrect.**
A 4-fold difference corresponds to a ΔCt of 2 cycles ($2^2 = 4$). The plot shows the black curves crossing the threshold much later than 2 cycles after the red curves—about 6 cycles later. Thus, 4-fold severely underestimates the true difference in starting template.

**B. 8-fold – Incorrect.**
An 8-fold difference corresponds to a ΔCt of 3 cycles ($2^3 = 8$). The visual gap between the red and black threshold crossings is larger than 3 cycles; the black curves cross roughly at cycle 28–29 compared with 22–23 for the red curves. Treating this as only an 8-fold difference ignores about half of the Ct separation and still underestimates the change in template amount.

**C. 64-fold – Correct.**
The threshold-crossing difference is approximately:

$$\Delta Ct \approx 28 - 22 \approx 6\ cycles.$$

Under ideal conditions, each PCR cycle doubles the amount of product, so the fold-difference in starting quantity is:

$$2^{\Delta Ct} \approx 2^6 = 64.$$

Because the red and black curves have similar slopes and plateau levels, it is reasonable to assume similar amplification efficiencies, making the ΔCt method appropriate. Therefore, the red group has roughly 64 times more initial template than the black group.

---

**D. 256-fold – Incorrect.**
A 256-fold difference would require a ΔCt of 8 cycles ($2^8 = 256$). The actual separation between the red and black curves at the threshold is closer to 6 cycles, not 8. Choosing 256-fold would overestimate the Ct gap and exaggerate the difference in starting template amount relative to what is shown in the figure.

**Q2 – IMAC Purification of His-tagged SWP5 (SDS–PAGE)**

The image shows an SDS–PAGE analysis of His-tagged SWP5 after immobilized metal affinity chromatography (IMAC). Panels (a) and (b) are replicate gels (independent runs) of the same IMAC workflow.
In both panels:

- **Lane 1**: Molecular-weight marker.
- **Lane 2**: 100 mM imidazole elution fraction from a Ni–NTA column.

**Which statement best describes the purification outcome shown in the image?**
A. SWP5 (~44.5 kDa) is present mainly in the marker lane; the elution lane does not show a species near the expected MW, indicating poor capture by IMAC.
B. The elution lane is **consistent with** successful enrichment of SWP5 near the expected MW (~44.5 kDa), with limited co-eluting species.
C. The elution lane is dominated by lower-molecular-weight species, consistent with extensive degradation and/or poor enrichment of intact SWP5.
D. The elution lane is essentially blank, consistent with failure to elute under 100 mM imidazole.

**Correct Answer**
**B. The elution lane is consistent with successful enrichment of SWP5 near the expected MW (~44.5 kDa), with limited co-eluting species.**

**Detailed Option Analysis**
**A – Incorrect.**
Option A is incorrect because Lane 1 is a molecular-weight marker and does not represent where SWP5 is detected in the sample. It only provides reference band positions. The claim that SWP5 is mainly in the marker lane is, therefore, conceptually wrong. In addition, the statement asserts that the elution lane lacks a species near ~44.5 kDa and concludes that IMAC failed to bind the target, which is not consistent with the visual evidence of a prominent signal in the elution lane at the expected molecular-weight region.

---

**B – Correct.**
Option B is correct because the expected molecular weight of SWP5 is approximately 44.5 kDa, and an IMAC elution that successfully enriches a His-tagged target should yield an elution lane whose dominant species aligns with the target's expected size. The pattern in the elution lane is consistent with a strong species near ~44.5 kDa, supporting the interpretation that SWP5 was captured by the Ni–NTA resin and released by 100 mM imidazole with relatively good enrichment compared with background proteins.

---

**C – Incorrect.**
Option C is incorrect because it characterizes the elution as being dominated by lower-molecular-weight species and lacking a prominent species near ~44.5 kDa, which would imply extensive degradation and/or poor enrichment of intact SWP5. That interpretation does not align with the observed outcome, where the elution lane shows its most prominent signal in the region corresponding to the expected size of SWP5. While minor lower-molecular-weight species could occur, they are not sufficient here to justify the conclusion that degradation or poor enrichment is the defining feature of the elution.

---

**D – Incorrect.**
Option D is incorrect because it requires the elution lane to be essentially blank and attributes this to a mechanistic failure, such as loss of metal ions from the resin, preventing elution. The image does not show an absence of eluted material in lane 2, so the premise of the option is not met. Moreover, invoking metal stripping as the explanation is an unnecessary mechanistic leap in the absence of supporting controls (for example, comparing flow-through, wash fractions, or a higher-imidazole elution series).

**Q9 –Inoculum Readiness: Adherent Mammalian Cell Culture Condition (Phase-Contrast Image)**

You are monitoring an adherent mammalian cell culture used as the inoculum for a fed-batch production run. The phase-contrast image shows an adherent mammalian cell culture near the end of its growth phase.

**Which statement best describes the condition of this culture based on the image?**
A. The culture is a uniformly healthy, well-spread monolayer and is ideal for use as an inoculum.
B. The culture is in very early growth; cells are sparse but healthy, so it should be expanded further before inoculation.
C. Many cells appear rounded or partially detached, with uneven density and dark clumps of debris, suggesting stress or possible contamination; this culture is not ideal for inoculation.
D. The culture shows no adherent cells at all, indicating that the flask is empty and needs to be reseeded.

**Correct Answer**
**C. Many cells appear rounded or partially detached, with uneven density and dark clumps of debris, suggesting stress or possible contamination; this culture is not ideal for inoculation.**

**Detailed Option Analysis**
**A – Incorrect.**
A truly healthy adherent culture suitable for inoculation should show a uniform, well-spread monolayer with cells having characteristic morphology (e.g., spindle-shaped or polygonal) and minimal debris. In the image, however, there are multiple rounded or partially detached cells, non-uniform density, and visible dark clumps. This does not match the appearance of a uniformly healthy monolayer, so it is not "ideal" for inoculum.

**B – Incorrect.**
An early growth phase culture typically shows sparse but evenly distributed, well-adhered cells with clear morphology and little debris. Here, the issue is not just low density; rather, there are aggregates and rounded/detached cells, suggesting stress, degeneration, or contamination, not a simple "still growing" state. Saying it should only be "expanded further" ignores these abnormal features.

**C – Correct.**
The image shows several hallmarks of a compromised culture:

- Many cells are rounded or partially detached instead of uniformly spread.
- There are dark clumps or aggregates
- Cell density appears uneven, with patches of sparsity and localized clusters.

These features are consistent with cell stress, reduced viability, and/or contamination, making the culture unsuitable as a clean, reliable inoculum for a production run. Therefore, option C correctly interprets the image and its implications.

**D – Incorrect.**
The flask is clearly not empty—there are many visible cells and structures. The problem is the quality of the culture, not the absence of cells. Thus, describing the flask as empty and in need of re-seeding is inconsistent with what is shown.

Across the pilot multimodal benchmark, several frontier models achieved strong performance, with Gemini-2.5-Flash, Gemini-2.5-Pro, Kimi-K2.5, and Qwen-3.5-9B each answered 8 of 10 questions correctly (80% accuracy; **Fig. 5A**). However, given the small pilot sample, these values are reported descriptively and should not be interpreted as precise estimates of relative model performance. These results suggest that tested multimodal LLMs can correctly interpret some experimental images, plots, and experimental readouts when the task is framed as a structured multiple-choice question. Notably, the more compact Qwen-3.5-9B achieved the same accuracy as the top-performing cloud-scale models. Unexpectedly, even some highly compact models, including Qwen-3.5-0.8B and DeepSeek-VL2-Tiny, outperformed or matched several larger models on this small pilot set (**Fig. 5A**). Again, this observation should be interpreted cautiously because the benchmark currently contains only 10 questions, making accuracy highly sensitive to individual items. Therefore, although these preliminary results suggest that edge-deployable multimodal models may have practical utility for selected tasks, a larger and more diverse set of questions will be required to robustly distinguish model capabilities.

We also evaluated the semantic similarity between model-generated and expert-authored explanations for the multimodal questions. In contrast to the MCQ benchmark, where many models showed broadly similar explanation-text similarity distributions, the multimodal benchmark showed clearer model-dependent variation (**Fig. 5B**). Cloud-scale models generally exhibited more stable explanation-text similarity scores, indicating more consistent semantic agreement between their post-hoc explanations and expert-authored interpretations. Qwen-3.5-9B achieved high answer accuracy but showed a broader distribution of explanation-text similarity, demonstrating that models with similar answer-selection performance can differ in how closely their generated explanations align with expert-authored explanations. Similar contrasts were observed among other edge-deployable and cloud-scale models. These results distinguish answer-selection accuracy from the semantic alignment of subsequently generated explanation texts, rather than directly measuring differences in the latent reasoning processes used to reach those answers.

Finally, we analyzed multimodal performance in relation to estimated model scale and deployability (**Fig. 5C**). The quadrant analysis shows that several frontier cloud-scale models occupy the high-accuracy region, consistent with their stronger multimodal reasoning capacity. However, the placement of Qwen-3.5-9B and other locally deployable models demonstrates that strong performance on the current multimodal questions is not restricted to cloud-accessed models. Together, these results provide preliminary evidence that multimodal LLMs may assist in interpreting figures, experimental reports, and manuscript-level visual data used in BE research. Nevertheless, broader benchmarking across additional subfields, image modalities, and experimental contexts will be necessary before assessing whether these capabilities can translate to potential downstream applications such as peer-review support, autonomous experimental interpretation, or laboratory automation.

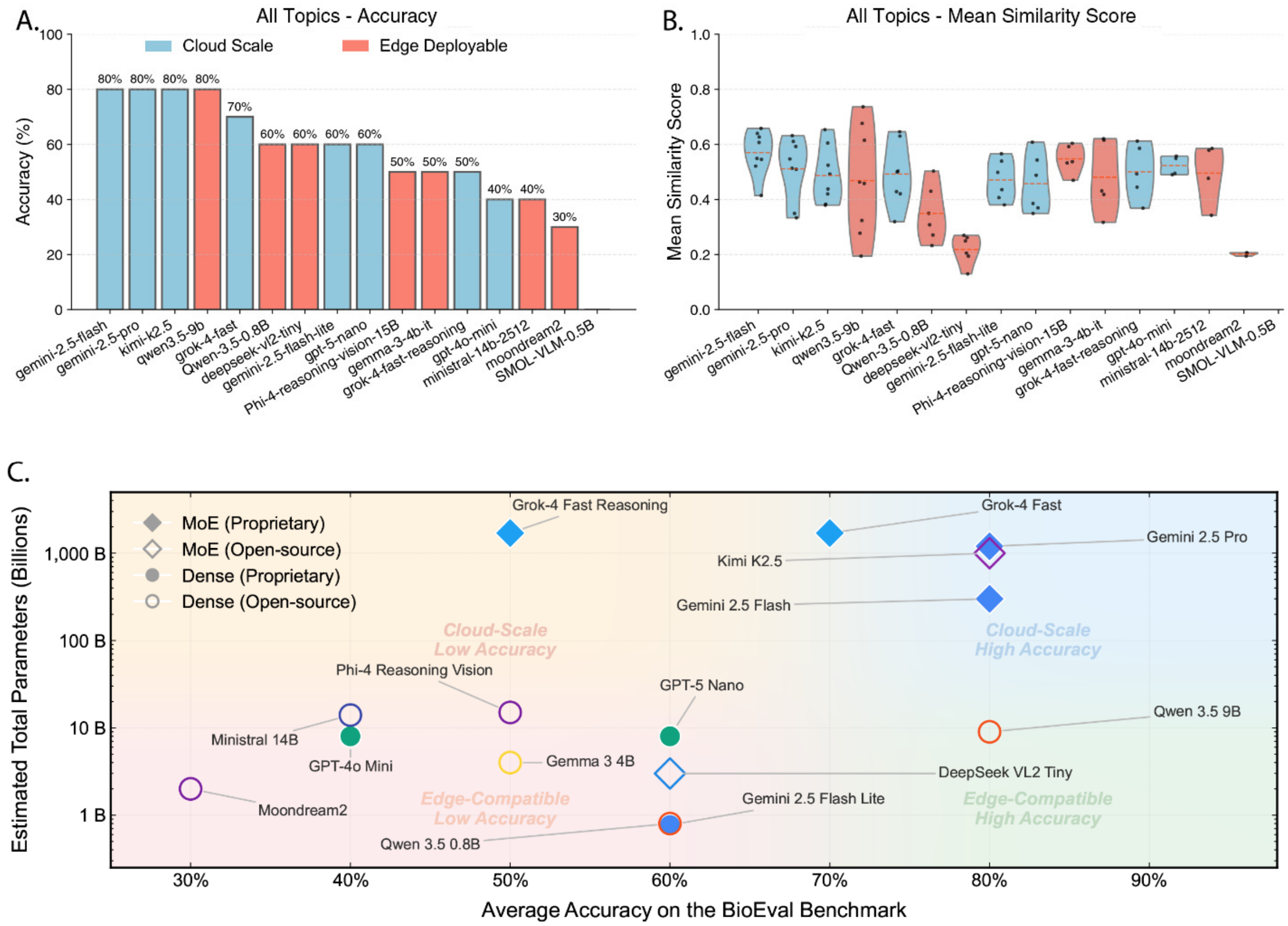

**Fig. 5** ***Evaluation of multimodal reasoning performance in BioEVAL.*** (**A**) Overall accuracy of cloud-scale and edge-deployable LLMs on the pilot ***BioEVAL*** multimodal reasoning question set. Blue bars indicate cloud-scale models, and pink bars indicate edge-deployable models. The current pilot set contains 10 expert-curated questions that require joint interpretation of experimental images and domain-specific BE reasoning. (**B**) Distribution of expert–model explanation-text similarity scores. Models were prompted after answer selection to generate explanations for all four options, and each generated explanation was compared with the corresponding expert-authored explanation using the semantic similarity metric. The four option-level scores were averaged per question. This metric assesses semantic agreement between post-hoc explanation texts. (**C**) Relationship between multimodal reasoning accuracy and estimated total parameter count. Models are organized by deployability and performance, with cloud-scale and edge-compatible regions indicated by color gradients. Because the current multimodal benchmark contains only 10 pilot questions, these results should be interpreted as preliminary and primarily used to guide expansion of the multimodal ***BioEVAL*** dataset. In addition, rankings in this Figure reflect observed benchmark accuracy and are descriptive; uncertainty intervals and formal pairwise significance testing were not performed in the current analysis.

## *III-5.* Correctness-Pattern Agreement and Ensemble Performance across LLMs

LLMs may show similar or distinct patterns of success and failure across benchmark questions. To characterize this structure, we evaluated whether models tended to answer the same items correctly or incorrectly and whether combining models through majority-vote ensembles could improve ***BioEVAL*** performance beyond that of the best individual model. Ensemble gains are expected when constituent models are both accurate and diverse in their correctness patterns: if models fail on different questions, majority voting can correct errors made by individual models.

Conversely, if high-performing models tend to fail on the same questions, simple majority voting may provide only limited benefit.

To quantify pairwise correctness-pattern agreement, we computed Cohen's κ values between binary correctness vectors for the 21 evaluated models on the MCQ benchmark (**Fig. 6A**). The analysis was performed on the 332 questions that exhibited cross-model variation, excluding 27 invariant items that were either answered correctly (26 questions) or missed (1 questions) by all models. Hierarchical clustering based on κ-distance revealed three main groups of models with similar correctness patterns, along with several unclustered models (**Fig. 6A**).

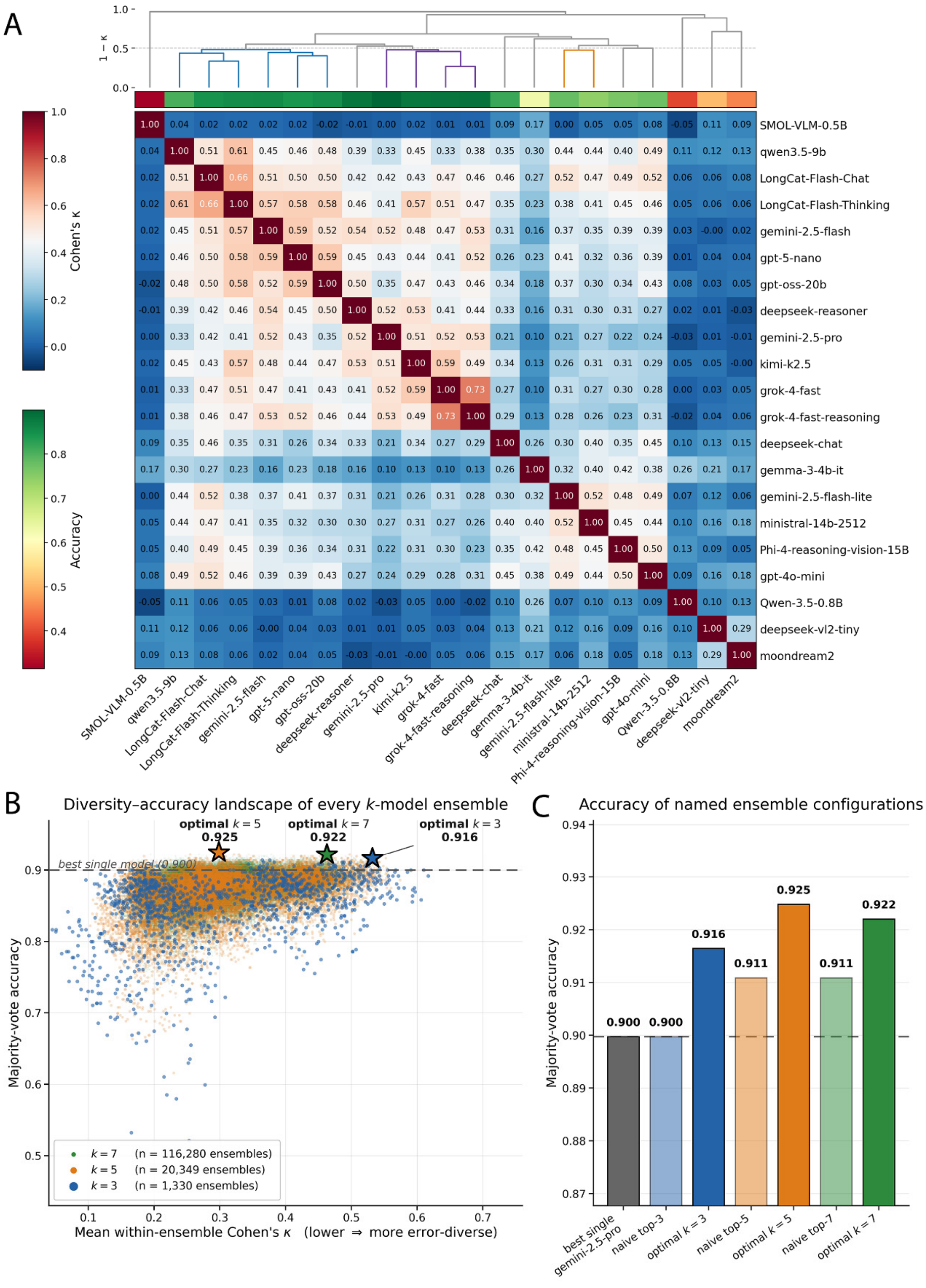

**Fig. 6** ***Correctness-Pattern Agreement and Ensemble Performance across LLMs.*** (**A**) Correctness-pattern agreement between 21 LLMs on the MCQ benchmark, quantified as pairwise Cohen's κ on binary correctness vectors. κ is computed on the 332 questions exhibiting cross-model variation; 27 invariant items (26 universally correct, 1 universally missed) carry no inter-rater signal and are excluded. Hierarchical clustering on κ-distance $d = 1 - κ$ with average linkage, cut at $d = 0.5$, partitions the 21 models into three coherent groups with substantive within-cluster

agreement, plus a tail of effectively unclustered models. (**B**) A scatter plot showing ensemble accuracy as a function of error diversity (Cohen's $\kappa$) for 3-, 5-, and 7-model ensembles, with ties broken by the highest-accuracy ensemble member. (**C**) Accuracy of the top-k ensembles. Naive top-k ensembles contain the models with the best overall accuracy as shown in **Fig. 2**, with the naive top-3, -5, -7 ensembles respectively comprising (Gemini-2.5-pro, grok-4-fast-reasoning, grok-4-fast), (Gemini-2.5-pro, grok-4-fast-reasoning, grok-4-fast, kimi-k2.5, deepseek-reasoner) and (Gemini-2.5-pro, grok-4-fast-reasoning, grok-4-fast, kimi-k2.5, deepseek-reasoner, Gemini-2.5-flash, LongCat-Flash-Thinking). The in-sample best 3-, 5-, and 7-model ensembles identified by exhaustive search on the same 359 benchmark questions are (Gemini-2.5-pro, deepseek-reasoner, kimi-k2.5), (Gemini-2.5-pro, grok-4-fast, deepseek-reasoner, gpt-oss-20b, qwen-3.5-0.8B), (Gemini-2.5-pro, grok-4-fast-reasoning, grok-4-fast, deepseek-reasoner, kimi-k2.5, LongCat-Flash-Chat, gpt-4o-mini). Additionally, these exhaustively selected ensemble accuracies are descriptive in-sample results and were not evaluated on a held-out question set.

The strongest reasoning models, including Gemini-2.5-Pro, Grok-4-Fast-Reasoning, Grok-4-Fast, Kimi-K2.5, and DeepSeek-Reasoner, formed a top-reasoner cluster with substantial within-cluster correctness agreement (**Fig. 6A**). This indicates that these models not only achieved similar high accuracy but also tended to succeed and fail on many of the same questions. Such overlap in correctness patterns may help explain why simple ensembles of the top-performing models provided only modest gains over the best single model (**Figs. 6B** & **C**).

A second cluster contained several fast or non-reasoning models, including Gemini-2.5-Flash, GPT-5-Nano, GPT-oss-20B, Qwen-3.5-9B, and the LongCat variants (**Fig. 6A**). Within this group, Qwen-3.5-9B and Gemini-2.5-Flash showed particularly strong similarity to the LongCat models, suggesting overlapping correctness patterns. A third mid-tier cluster grouped smaller but competent generalist models, including DeepSeek-Chat, Gemini-2.5-Flash-Lite, Ministral-14B-2512, Phi-4-Reasoning-Vision-15B, and GPT-4o-mini. The lowest-accuracy unclustered models in **Fig. 6A** showed near-zero $\kappa$ with competent models, indicating relatively distinct correctness patterns, but their low accuracy limited their usefulness in majority-vote ensembles.

We then exhaustively evaluated all possible 3-, 5-, and 7-model ensembles using majority voting on the 359-question MCQ benchmark (**Fig. 6B** & **C**). The best single model achieved an observed accuracy of 0.900. Naive top-k ensembles, formed by selecting the individually highest-accuracy models, produced little improvement: the naive top-3 ensemble matched the best single-model accuracy at 0.900, whereas the naive top-5 and top-7 ensembles reached 0.911. Exhaustive search identified in-sample best ensembles with observed accuracies of 0.916, 0.925, and 0.922 for ensemble sizes of 3, 5, and 7, respectively. Because ensemble composition was selected and evaluated on the same benchmark questions, these values represent in-sample, oracle-selected performance and should not be interpreted as estimates of improvement on unseen questions.

The in-sample best ensembles were not always composed only of the highest-performing individual models. For example, the best 5-model ensemble included Gemini-2.5-Pro, Grok-4-Fast, DeepSeek-Reasoner, GPT-oss-20B, and Qwen-3.5-0.8B, showing that lower-ranked or smaller models can sometimes contribute useful complementary votes when their errors differ from those of frontier models. However, the observed in-sample gain remained modest: the best five-model ensemble achieved 0.925 accuracy compared with 0.900 for the best single model, corresponding to approximately 9 additional correct responses among the 359 MCQs. Because this ensemble was selected on the same questions used for evaluation, the difference should be interpreted as an in-sample upper bound on the benefit obtainable through this exhaustive majority-vote search rather than evidence of improved generalization to unseen questions. Together, these results suggest that similar error patterns among high-performing models, may limit the benefit of the simple majority-vote ensemble approach to improve accuracy.

## IV. Discussion

In this study, we introduced ***BioEVAL*** as a global, multi-institutional benchmark for evaluating LLMs across PhD-level BE knowledge and tasks. Unlike many biomedical or general scientific benchmarks, ***BioEVAL*** focuses on research-relevant tasks that more closely resemble the daily work of experimental bioengineers: interpreting domain-specific MCQs, synthesizing scientific literature, and reasoning over experimental visual data. By integrating expert-curated MCQs, L-Syn tasks, and pilot multimodal reasoning problems across major BE subfields, ***BioEVAL*** provides both a quantitative leaderboard and a diagnostic map of where current models are strong, where they remain unreliable, and which directions should be prioritized for BE-oriented model development.

A central finding is that current frontier LLMs already demonstrate substantial BE domain knowledge. Several cloud-scale models achieved greater than 85% overall accuracy on the MCQ benchmark, with the leading model reaching 90% accuracy. This level of performance suggests that modern LLMs can correctly answer many PhD-level BE questions when the task is framed as structured multiple-choice reasoning. Importantly, strong performance was not limited to the largest cloud-scale systems. The edge-deployable GPT-oss-20B model achieved 85% accuracy, approaching the performance of several frontier cloud-scale models, while other compact systems, such as Qwen, also performed meaningfully above chance. These results are encouraging for BE applications because edge-deployable models could support local inference in privacy-sensitive research environments. These results motivate further evaluation of locally deployable models for potential applications including literature understanding, protocol generation, routine data interpretation, decision support, and AI-assisted laboratory automation; however, these downstream applications were not directly evaluated in ***BioEVAL*** and will require task-specific validation before deployment.

However, the results also indicate that high answer accuracy alone should not be interpreted as full experimental readiness. The separation between conceptual and experimental MCQs revealed that experimental questions were consistently more difficult for most models than conceptual questions. This distinction is critical for BE. Many real-world BE tasks are not simple knowledge-recall problems; they require integration of assay design, biological controls, material properties, device constraints, kinetic behavior, statistical uncertainty, and failure-mode analysis. A model that can recall a principle may still fail when asked to interpret an unexpected experimental outcome or select the most appropriate troubleshooting step.

Expert–model explanation-text similarity scores were often centered around intermediate values even for models with relatively high answer accuracy. Notably, this pattern was model-family-specific rather than uniform: the Qwen models achieved competitive answer accuracy on both the MCQ and multimodal benchmarks while showing lower mean explanation-text similarity and higher variance than some models with comparable accuracy. These results indicate that models with similar answer-selection performance can differ in the semantic alignment of their subsequently generated explanations with expert-authored explanations. However, because explanations were generated after answer selection, the metric should not be interpreted as a direct measure of the latent reasoning process that produced the answer. Moreover, the current automated metric has not been validated against independent human judgments of explanation quality. Sentence-embedding similarity may be influenced by explanation length, register, structure, and formatting compliance, particularly across model families and scales. Future work should therefore compare automated explanation-text similarity with expert human ratings on representative subsets and examine selected-option and correct-answer explanations separately.

Subfield-resolved performance further showed that LLM capability is not uniformly distributed across BE. Some areas, including *Diagnostics, Biosensing, and Bioelectronics, Neuroengineering and Neurobiology,* and *Immunoengineering*, showed consistently high MCQ performance across many models. In contrast, *Bioimaging, Biophotonics, and Optics*, *Biomaterials and Biomolecules*, *Genetics, Systems and Synthetic Biology*, and *Drug Delivery, Therapy, and Nanomedicine* showed lower performance across most models. These differences should not be interpreted solely as a reflection of the volume of literature or subfield popularity. The supplementary analysis of BE subfield prevalence did not reveal a simple relationship between how frequently a subfield appears across academic institutions and how well LLMs perform in that subfield (**Fig. S1**). The present study does not establish the causes of these subfield-level differences. Potential contributors could include question composition and difficulty, contributor-specific item design, terminology and experimental complexity, differences in training-data representation, and the extent to which tasks depend on procedural laboratory knowledge; however, these factors were not independently measured or controlled in the current benchmark. Subfields with lower observed performance may nevertheless provide useful targets for future investigation to determine whether targeted dataset curation, retrieval-augmented workflows, fine-tuning, or other model-development approaches can improve performance.

The L-Syn benchmark revealed a complementary aspect of model capability. Several models achieved similarity scores above 0.70, indicating that LLMs can extract and reorganize substantial information from abstract-withheld full-text articles (capped at 50,000 tokens as described in Methods V-7) and generate plausible publication-style summaries. But, importantly, embedding-based similarity quantifies semantic agreement with the reference abstract and does not establish factual accuracy. Sentence embeddings are largely insensitive to numerical magnitude and to the direction of quantitative claims, and the natural-language-inference term penalizes explicit contradiction rather than fabrication. For instance, an assertion absent from the reference is scored as neutral rather than contradictory. A generated abstract containing incorrect values, inverted effects, or unsupported claims may therefore still receive a high similarity score. These scores should be interpreted as measures of semantic coverage and topical fidelity rather than as evidence that model-generated abstracts are free of hallucination. The same caveat applies to the automated expert–model explanation-text similarity used in the MCQ and MRQ benchmarks (Methods V-6, V-8).

Also, even the highest-scoring output remained imperfect approximations of the original author-written abstracts. Scientific abstracts are not merely compressed summaries; they reflect expert judgment about novelty, quantitative emphasis, mechanistic interpretation, and the significance of the central findings. Models may therefore recover much of the information contained in an article while missing its most important experimental insight or mis-prioritizing background relative to the primary contribution. These results support the near-term use of LLMs for rapid literature comprehension and draft summarization, but do not yet support replacing the scientific judgment required for expert-level synthesis. Moreover, because the pretraining corpora of several evaluated models are undisclosed, prior exposure to some included articles or abstracts cannot be excluded; abstract withholding prevents direct access during evaluation but does not eliminate potential pretraining contamination. Although article titles, DOIs, and bibliographic metadata stored as separate dataset fields were not supplied to models, residual identifying information embedded within some prepared full-text inputs cannot be excluded; consequently, the L-Syn task does not fully isolate de novo synthesis from possible paper recognition. Future evaluations should extend beyond semantic similarity to assess novelty capture, quantitative fidelity, mechanistic accuracy, and appropriate prioritization of major findings.

The pilot multimodal benchmark highlights another major frontier for BE-oriented AI. Experimental BE is inherently visual and multimodal: researchers routinely interpret microscopy images, gels, Western blots, growth curves, fluorescence readouts, flow cytometry plots, chromatograms, device schematics, and other instrument-generated outputs. In the pilot multimodal evaluation, several models achieved 80% accuracy, suggesting that current visual/multimodal LLMs can interpret a meaningful fraction of experimental images and plots when the task is well structured. These preliminary results motivate future evaluation of multimodal LLMs for potential applications such as automated quality control and laboratory robotics (which were not directly tested in the present study). The explanation-text similarity analysis further indicates that models with comparable visual answer-selection accuracy can differ in the semantic alignment of their post-hoc explanations with the expert-authored. Thus, expanding and testing the multimodal benchmark should be a priority. A substantially larger MRQ dataset, ideally containing hundreds of questions across diverse BE subfields, image modalities, and perhaps even audio modalities (such as machine noises), will be necessary to determine whether current models can reliably interpret and draw useful conclusions from experimental evidence.

The error-pattern and ensemble analyses provide an additional caution for deployment. Although ensemble voting is often assumed to improve model reliability, even the exhaustively selected in-sample best ensemble showed only a modest observed gain, increasing accuracy from 0.900 to 0.925 (approximately 9 additional correct responses among 359 MCQs). Because ensemble composition was optimized and evaluated on the same questions, this result represents an in-sample upper bound. This limited improvement is likely because most of the tested models are not independent; they share correlated error patterns. Future ensemble strategies may need to prioritize complementary reasoning behavior rather than simply aggregating models based on their accuracy. Future evaluations should use held-out questions or nested resampling to separately select and evaluate ensemble composition.

The hardest-item analysis revealed the tendency of models to have collective blind spots, which is likely driven by shared training corpora. The hardest-item analysis additionally highlights the value of model-response patterns for benchmark quality control: unusually low cross-model accuracy or strong convergence on a common non-keyed answer may flag questions that warrant renewed expert review.

This study has several limitations. First, although ***BioEVAL*** was constructed by domain experts and underwent group-level expert review and centralized quality control, the current version remains limited by available community resources. Future iterations should include more systematic multi-stage validation, broader participation from additional BE subfields, and explicit human baseline comparisons, including undergraduate, graduate student, postdoctoral, and faculty-level performance.

Second, the number of questions per subfield should be expanded. A more balanced benchmark with at least 100-200 MCQs per subfield would better support statistically robust comparisons, reduce sensitivity to individual question difficulty, and allow more detailed analysis of conceptual versus experimental reasoning. Because each subfield's items were contributed by a small number of research groups, subfield identity is partially confounded with contributor identity.

Third, the model rankings reported in the current study are based on observed benchmark accuracies without confidence intervals or formal paired comparisons between models. Accordingly, small differences in accuracy, particularly among closely ranked models, should not be interpreted as statistically significant differences in capability. This limitation is particularly important for the pilot multimodal benchmark, which contains only 10 questions, such that each individual response changes model accuracy by 10 percentage points. In addition, each model–question pair was evaluated once using deterministic decoding where supported, so the present study does not quantify run-to-run variability arising from stochastic inference, model-serving infrastructure, or API/backend behavior. Future evaluations should incorporate question-level uncertainty estimates, paired model comparisons, held-out or resampled evaluation of ensemble selection, repeated inference where appropriate, and larger per-subfield and multimodal evaluation sets.

Fourth, the L-Syn benchmark was evaluated under a standardized input-length constraint to accommodate models with smaller context windows, particularly locally deployable models. Article inputs exceeding approximately 50,000 tokens were therefore truncated, whereas shorter articles were provided without truncation. Consequently, some L-Syn tasks were evaluated using only a portion of the abstract-withheld article rather than the complete article text, though all models were exposed to the same truncated text irrespective of the size of their context window. Future evaluations could incorporate models with longer context windows or analyze subsets of articles that can be provided in full across all evaluated models.

Fifth, the multimodal benchmark remains a small feasibility pilot and should be expanded substantially, ideally to hundreds of questions spanning microscopy, gels, flow cytometry, spectroscopy, sequencing traces, omics visualizations, device schematics, and quantitative experimental plots.

Sixth, this study primarily evaluates general-purpose foundation and multimodal models. Domain-specific or fine-tuned models, including biomedical, biological, and scientific LLMs, should be evaluated in future work to determine whether specialized training improves BE reasoning beyond general model scaling.

Seventh, curation and access remain centralized. Benchmark items were contributed by participating groups, but answer-key adjudication, rationale revision, and final inclusion decisions were made centrally by the coordinating team, which also administers access to the full item set through the evaluation API. This arrangement protects benchmark integrity against training-data leakage during the initial release but concentrates editorial authority narrowly. Establishing transparent community governance, including a standing multi-institution review panel with defined procedures for adjudicating contested items and documented criteria for granting data access, is therefore a priority for subsequent releases.

Finally, because LLMs evolve rapidly, these results represent a snapshot of capabilities between Nov. 2025 and Feb. 2026, and we deliberately refrain from extrapolating them to subsequent releases. However, our data indicate that ***BioEVAL*** performance may not simply reflect scale or recency: GPT-oss-20B achieved higher observed accuracy than several cloud-accessed systems, while compact models matched frontier systems on selected multimodal items.

To support broader community participation, future versions of ***BioEVAL*** will be structured around two participation pathways. First, model developers, academic groups, or companies interested in evaluating new models will be able to test them using a standardized ***BioEVAL*** evaluation workflow, with consistent prompts and scoring metrics for accuracy, explanation-text similarity, literature-synthesis similarity, and multimodal reasoning performance. To reduce benchmark leakage into LLM training data, full benchmark items and answer keys may be distributed through individual requests. Second, BE research groups interested in contributing to the benchmark will be able to submit new MCQs, L-Syn articles, and MRQs using standardized templates. Submitted items would undergo expert review and independent cross-group assessment within the same subfield, with disagreements resolved through predefined revision or exclusion procedures before inclusion in subsequent benchmark releases. Through this structure, ***BioEVAL*** can move beyond a static benchmark and serve as an evolving community initiative for standardized evaluation of LLMs tuned to BE-specific challenges. As the dataset expands, the initiative could support open model comparison, controlled leaderboard updates, subfield-specific benchmark extensions, and collaborative development of higher-quality expert-curated evaluation items. Ultimately, advancing AI-assisted BE will require better benchmarks comprising richer expert-curated datasets (both text and multimodal formats), and sustained collaboration between LLM developers introducing larger and uniquely tuned models and experimental BE communities.

## V. Method

### *V-1. BioEVAL* Study Design

***BioEVAL*** was designed as a global, multi-institutional benchmarking framework to evaluate the capabilities and limitations of large language models (LLMs) in bioengineering (BE). The study was initiated to address the lack of BE-specific benchmarks that assess not only factual recall, but also experimental reasoning, literature comprehension, and multimodal interpretation in research-relevant contexts. The overall objectives of ***BioEVAL*** were to: (1) construct a rigorous, expert-curated BE benchmark dataset; (2) systematically evaluate the performance of current cloud-scale and edge-deployable LLMs; (3) identify model strengths, weaknesses, and subfield-specific failure modes; (4) establish standardized evaluation procedures for BE-oriented LLM benchmarking; and (5) provide a benchmarking foundation for future evaluation and potential implementation of LLMs in BE research, AI-assisted experimental workflows, and laboratory automation.

The ***BioEVAL*** benchmark was organized around three complementary evaluation categories. The first category consisted of multiple-choice questions (MCQs) designed to assess BE domain knowledge, conceptual understanding, and experimental reasoning. Each MCQ included four answer options, one correct answer, and expert-written rationales explaining why each option was correct or incorrect. The second category consisted of literature synthesis (L-Syn) tasks, in which models were provided with full-text research articles (capped at 50,000 tokens as described in Methods V-7) with abstracts withheld and prompted to generate publication-style abstracts. These tasks were designed to evaluate whether LLMs could extract, prioritize, and synthesize key scientific findings from primary literature. The third category consisted of multimodal reasoning questions that required models to interpret experimental images, plots, or graphical results alongside domain-specific BE knowledge.

***BioEVAL*** was constructed as a community-driven benchmark spanning 11 major BE subfields and a diversified catalog of contributed questions that are beyond the scope of these subfields. Participating research groups contributed expert-curated items reflecting research topics and experimental workflows from their respective domains. To improve benchmark quality, contributed items underwent expert review and centralized quality control prior to model evaluation. This process was intended to reduce ambiguity, remove poorly supported questions, and ensure that each benchmark item had a clearly defined correct answer or reference output.

The benchmark resulting from authoring-group review and centralized quality control was used to evaluate a diverse set of LLMs, including frontier cloud-scale models, reasoning-focused models, visual/multimodal models, and locally deployable models, all available in Feb. 2026. Model performance was assessed using task-specific metrics: accuracy and expert–model explanation-text similarity for MCQs, semantic precision and coverage for L-Syn tasks, and accuracy and explanation-text similarity for multimodal reasoning questions. Downstream analyses included overall model ranking, cloud-scale versus edge-deployable model comparison, subfield-resolved performance analysis, explanation-text similarity analysis, error-pattern similarity analysis, and ensemble-performance evaluation.

### *V-2.* Selection of BE Subfields and Participating Expert Groups

Because bioengineering (BE) spans a broad and heterogeneous research landscape, we first performed a structured survey to define representative BE subfields for benchmark construction. This survey was based on the assumption that major global research institutions are among the most active contributors to academic BE research and therefore provide a reasonable approximation of contemporary subfield distribution.

First, a list of leading global research institutions was assembled using the 2025 U.S. News and QS global university rankings. Duplicate institutions appearing across both rankings were identified and merged, resulting in an initial shortlist of approximately 120 institutions worldwide. Each institution was then manually reviewed through publicly available departmental websites. Institutions without identifiable BE departments, biomedical engineering programs, or closely related BE research activities were excluded from downstream subfield-frequency analysis.

For each remaining institution, the BE-related research subfields represented were recorded in a structured spreadsheet. Because different institutions used heterogeneous terminology to describe overlapping research areas, subfield labels were subsequently standardized using a Python-based semantic grouping workflow. Closely related terms and overlapping research descriptors were merged to generate a consolidated list of 22 categories, consisting of 21 major BE-related subfields and one additional category for minority or less frequently represented subfields, as shown in **Fig. S1**.

Following the subfield-frequency analysis, participating research groups were recruited through existing academic networks, with outreach prioritized from higher-frequency BE subfields toward lower-frequency subfields. To support future cross-group validation during the second phase of benchmark construction, only subfields with at least two participating research groups were included in the initial ***BioEVAL*** dataset. This criterion was intended to enable that each included subfield could support independent expert review and consensus-based validation of contributed benchmark items later on.

Based on this process, 11 BE subfields and a set of uncategorized questions were selected for the first ***BioEVAL*** benchmark release, as summarized in **Fig. 3A**. Participating groups from universities worldwide, as listed in **Fig. 1A**, contributed benchmark items within their areas of expertise, including MCQs, L-Syn tasks, and multimodal reasoning questions. The resulting subfield structure was therefore informed by both the global academic prevalence survey and the practical feasibility for multi-group expert validation within each included subfield.

### *V-3.* Benchmark Dataset Construction

The ***BioEVAL*** benchmark was constructed to evaluate complementary dimensions of LLM performance across BE, including domain knowledge, experimental reasoning, literature synthesis, and multimodal interpretation. Three dataset categories were generated: multiple-choice questions (MCQs), literature synthesis (L-Syn) tasks, and multimodal reasoning questions (MRQs). For each category, participating research groups contributed expert-curated materials within their respective BE subfields, and all submitted items were organized into structured tabular formats for downstream model evaluation and analysis.

For the MCQ dataset, each participating research group was asked to identify at least one domain expert, for example, a PhD student, postdoctoral researcher, faculty member, or principal investigator, to generate approximately 20 PhD-level MCQs within the group's area of expertise (where 'PhD-level' denotes research-oriented questions authored by PhD-level domain experts and intended to reflect graduate research knowledge and experimental reasoning rather than a calibrated human-performance threshold). Each MCQ was written in a four-option format with one unambiguous correct answer. Contributors were also asked to provide a rationale for each answer option, explaining why the correct option was correct and why the remaining options were incorrect. Rationale length was generally limited to fewer than 200 words per option. Experts were encouraged to write rationales of at least 50 words when feasible to support meaningful semantic similarity scoring between LLM-generated and expert-written rationales. Shorter rationales were permitted for intrinsically straightforward options, numerical calculation-based questions, or cases in which a concise explanation was sufficient. After collection, all MCQs, answer choices, correct answers, expert rationales, and subfield labels were merged into a unified Pandas DataFrame. Each MCQ was additionally annotated as either experimental or conceptual. Experimental questions were defined as items requiring practical or literature-informed knowledge of experimental design, execution, analysis, interpretation, or troubleshooting, typically involving reasoning about research scenarios rather than direct factual recall. Conceptual questions primarily assessed established principles, properties, mechanisms, functions, or definitions that could be answered without substantial experimental-context reasoning. This annotation enabled downstream comparison between research-oriented experimental reasoning and primarily conceptual knowledge. The MCQ dataset used for the reported model evaluations comprised 380 unique entries (359 items retained after audit).

For construction of the L-Syn dataset, participating research groups were asked to identify current or ongoing research questions relevant to their respective BE subfields and to nominate closely related primary research articles addressing similar scientific problems or experimental themes. Contributors were encouraged to select approximately five highly relevant articles per research question, with preference given to recent literature (ideally published within the preceding five years), although the number of contributed articles varied across research groups and subfields. Only traceable primary research articles with a DOI were included; case studies, technical reports, websites, and other non-primary sources were excluded; and research teams are requested to provide primary research articles as substitution. This expert-guided selection strategy was intended to capture literature representative of active research directions across the BE community rather than textbook-like or historically established topics. For each article, the original abstract and a corresponding abstract-withheld full-text version were prepared. Article titles and DOIs were retained as separate dataset metadata fields for article tracking but were not provided to the evaluated models. Model inputs consisted only of the prepared abstract-withheld full text, which retained the main textual content and figure captions but excluded figure image files, enabling identical text-based evaluation of text-only and multimodal LLMs. The prepared full texts were not systematically screened for residual identifying information embedded within the source text, such as author names, affiliations, journal information, or DOI strings. The original author-written abstracts were retained separately as reference outputs for semantic similarity evaluation. Article metadata, subfield and research-question labels, abstracts, and abstract-withheld full texts were consolidated into a structured Pandas DataFrame. The final L-Syn dataset comprised 218 unique primary research articles.

For the pilot MRQ dataset, an initial set of 10 multimodal reasoning questions was generated. Each MRQ was designed to require joint interpretation of an experimental image, plot, or graphical readout together with BE domain knowledge. Similar to the MCQ dataset, each MRQ included four answer options, one unambiguous correct answer, and expert-written rationales for each option. The pilot dataset included representative experimental data types such as qPCR amplification curves, SDS–PAGE gels, bacterial growth curves, fed-batch fermentation profiles, mycoplasma PCR quality-control gels, Sanger sequencing chromatograms, adherent mammalian cell culture images, and enzyme-kinetics plots. Because this multimodal dataset was generated as an initial pilot set, it was used primarily to assess feasibility and provide preliminary insight into model performance on BE visual reasoning tasks. All MRQ prompts, image references, answer options, correct answers, rationales, and metadata were organized into a structured tabular format for downstream evaluation.

### *V-4.* Dataset Review and Quality Control

For the current preprint-stage release, benchmark quality control before model evaluation consisted of two primary stages: review within each contributing research group, followed by centralized review by the coordinating team. The broader BioEVAL validation framework is intended to incorporate independent cross-group review by multiple domain experts within each BE subfield; however, this full multi-group validation procedure was not completed across the entire current dataset before model evaluation. A separate targeted cross-group audit of selected MCQs was conducted after model evaluation, as described in V-4.1.

For the MCQ and L-Syn datasets, primary quality control was first performed within each contributing research group. Because ***BioEVAL*** recruited domain experts through traceable academic research groups rather than anonymous online contributors, submitted items were linked to identifiable research teams with relevant expertise and accountability. Each group's principal investigator reviewed and revised the submitted materials prior to central aggregation. This review focused on whether MCQs had a single unambiguous correct answer, whether distractor options were plausible but clearly incorrect, whether expert rationales accurately explain each option, and whether L-Syn article selections were relevant to the stated research questions.

After group-level review, the aggregated MCQ and L-Syn datasets underwent centralized quality control. This review included checking formatting consistency, subfield labels, conceptual versus experimental/literature-backed annotations, answer-key integrity, rationale completeness, duplicated or overlapping items, and consistency between submitted article metadata, abstracts, and abstract-withheld full-text files. Most dataset entries were revised through at least one to two rounds of iteration to improve clarity, remove ambiguity, and standardize the format for downstream model evaluation.

The pilot multimodal reasoning question (MRQ) dataset underwent a separate expert-review workflow by two research groups. The MRQ author then revised the dataset through iterative feedback.

***V-4.1 Post-evaluation cross-group audit of selected MCQ items***

Following evaluation of the 21 models on the 380-question MCQ benchmark, we performed a targeted cross-group quality audit to determine whether extreme model-response patterns could identify items warranting additional expert review. Two strata were selected from the model results: the 20 items answered correctly by the fewest models and the 20 items answered correctly by all 21 models. There were 31 questions answered correctly by all the models; 20 questions were selected as the first 20 entries when arranged by their question id, which was assigned by contributing lab (alphabetically arranged) and, within lab, by submission order. Similarly, the 20 least answered questions were obtained by aggregating the 18 questions answered correctly by 0, 1 or 2 models, with question ids used to select 2 of the 10 questions answered by just 3 models to complete the stratum. These strata represented response patterns potentially informative for benchmark quality control: near-universal disagreement with the benchmark key could reflect genuine difficulty but could also arise from ambiguous wording, a defensible alternative answer, or an answer-key problem, whereas universally answered items may provide limited discrimination among models. The two strata were pooled, randomized into a single 40-item list, and circulated without indication of stratum membership or model performance so that reviewers could not condition their judgments on either.

Reviewers were drawn from the participating research groups and were asked to vote only on items falling within their own expertise, recording one of three actions, including "retain", "revise", or "remove", together with a free-text justification. Responses were returned in whichever format each group preferred and were subsequently normalized onto this common three-action scale.

Blank entries and explicit abstentions ("not my expertise") were coded as non-votes rather than as tacit agreement, so that they reduced an item's vote count rather than inflating its apparent support. Two recurring response conventions required explicit coding rules: reviewers who returned only a selected answer for an item, without commentary, were coded as "retain"; and reviewers who recorded a combined "revise or remove" verdict without selecting between the two actions were retained as a separate, unassigned category and were not redistributed to either action.

Adjudication proceeded in two stages. An item was taken to have reached panel consensus when at least three experts had voted on it and at least two-thirds of those votes agreed on whether the item required any change. Within a needs-change consensus, an item was removed only where removal votes outweighed revision votes (unassigned "revise or remove" verdicts contributing one half to each side) so that an item was preferentially repaired rather than discarded. Items not meeting the consensus threshold were resolved by two pre-specified rules: an item with fewer than three votes and no dissenting vote was retained, and an item on which the panel was split (less than 2/3 votes for retention) was returned for revision. This audit was conducted across, but not outside, the contributing consortium, and reviewers may not be blinded to item authorship.

Audit verdicts were then applied to the reported results. Items adjudicated as requiring removal (n = 2) or revision (n = 19) were withheld from scoring, leaving 359 of the 380 evaluated questions. Because MCQ accuracy and the explanation-text similarity metric are computed per item, all reported MCQ analyses were recomputed over the retained subset from the original model responses; no model was queried again, and no response was regenerated. Items returned for revision are retained in the benchmark and are intended for reinstatement following expert revision in a subsequent release.

***V-4.2 Statistical analysis***

The distribution of adjudicated outcomes between the two strata was compared using two-sided Fisher's exact tests, applied separately to the proportion of items requiring any change (revise or remove) and to the proportion removed. Analyses were performed in Python (v3.14) and figures generated with Matplotlib (v3.11).

***V-4.3 Planned independent validation***

For the current preprint-stage benchmark, the dataset used for model evaluation underwent authoring-group review and centralized quality control before inference, whereas the 40-item cross-group audit described in V-4.1 was conducted post hoc. Items flagged by that audit were withheld from scoring, so the reported model results correspond to the 359 retained questions rather than the full 380-item set on which inference was performed. Future ***BioEVAL*** releases will implement broader independent cross-group validation before model evaluation. In this prospective workflow, contributed items will be reviewed by multiple domain experts, disagreements will be resolved through predefined revision or exclusion procedures, and the adjudicated benchmark will be frozen before model inference. This expanded validation framework is intended to improve benchmark robustness, reduce contributor-specific bias, and ensure that reported model results correspond to the finalized reviewed dataset.

## *V-5.* Model selection, inference settings, and prompting strategy

In the current preprint, we evaluated a panel of 21 contemporary LLMs. The selected models were characterized along four axes: (1) deployment mode and accessibility, distinguishing cloud-accessed models from locally deployable open-weight models that could be executed on consumer-grade GPUs; (2) reasoning capability, distinguishing models with explicit inference-time reasoning ("thinking") from non-reasoning models; (3) modality, distinguishing text-only from vision-capable multimodal models; and (4) architecture, including dense and mixture-of-experts (MoE) systems. The cloud-accessed models comprised Gemini-2.5-Pro, Gemini-2.5-Flash, Gemini-2.5-Flash-Lite, GPT-4o-mini, GPT-5-nano, Grok-4-Fast, Grok-4-Fast-Reasoning, DeepSeek-Chat, DeepSeek-Reasoner, Kimi-K2.5, and LongCat-Flash-Chat/-Thinking. The locally deployable open-weight group comprised GPT-oss-20B, Qwen-3.5-9B, Ministral-14B, Gemma-3-4B, Phi-4-reasoning-vision-15B, Qwen-3.5-0.8B, SMOL-VLM-0.5B, DeepSeek-VL2-Tiny, and Moondream2. This grouping was based on the deployment configuration used in the study rather than a strict parameter-count threshold. All 21 models were evaluated on the MCQ benchmark; the literature-synthesis (L-Syn) benchmark used the 20 text-capable models, and the pilot multimodal benchmark used the 16 vision-capable models (10 questions each).

All models were queried programmatically through a unified, OpenAI-compatible client interface, except for Gemini models, which were queried via the Google Gen AI interface. Proprietary, cloud-scale models were accessed via their original developer APIs (*e.g.*, OpenAI, Google Vertex AI for Gemini, xAI, DeepSeek, Moonshot, and LongCat) or via OpenRouter, whereas open-weight models were run locally on GPU workstations and on RunPod instances. Each prompt was presented as an independent, single-turn, zero-shot query—without in-context examples or conversation history. Tool use, including web search, was also disabled, so that the scores reflect each model's intrinsic capability to solve the presented problems. Decoding was deterministic (temperature = 0, where supported) for reproducibility, and option-wise token log-probabilities were recorded for the subset of APIs that expose them. The prompt responses were subsequently processed as described in the sections below. All model evaluations were conducted between November 2025 and February 2026; exact model version identifiers will be reported in the final manuscript.

Across tasks, prompting followed a common strategy. For the choice-based tasks with a fixed answer set (MCQs and multimodal questions), we used a two-stage design in which the model first selected an answer and then, in a separate call conditioned on that answer, generated structured explanations for all four options. The resulting explanations were therefore post-hoc outputs and were not interpreted as direct observations of the latent reasoning process that produced the answer. The literature-synthesis task instead used a single free-text generation call. In all cases, models were given a concise domain-expert role and instructed to generate structured outputs (a single option letter, a JSON rationale object, or an abstract-only response) to support automated, uniform scoring. Task-specific prompt wording and the corresponding scoring procedures are described in the following subsections.

***V-6.* MCQ prompts, accuracy and expert–model explanation-text similarity scoring**

Each MCQ was evaluated with the two-stage protocol described in V-5. In the first call, the model received the question and its four options and was instructed to return a single uppercase letter (A-D) with no additional tokens; the returned letter was taken as the model's prediction. In the second call, the model was asked to return a JSON object with one entry per option, each giving a correctness label and a brief mechanism, supporting evidence, and a one-to two-sentence free-text justification; this call was conditioned on the model's own selected answer from the previous call.

Answer accuracy was computed as the fraction of questions for which the predicted letter matched the expert answer key, with empty and malformed responses scored as incorrect. Accuracy was reported overall and by subfield (**Fig. 3A**), and separately for questions annotated as conceptual versus experimental (**Fig. 2C**, **Supplementary Figures**).

Because the explanation call returned a justification for every answer option, expert–model explanation-text similarity was computed option-wise. Each model-generated explanation was compared with the expert-authored explanation for the same option, independent of which option the model selected. Each MCQ therefore yielded four option-level similarity scores, which were averaged to obtain the per-question explanation-text similarity reported in **Fig. 2B**. When a model selected an incorrect option, explanations for all four options were still scored against their corresponding expert-authored explanations. Accordingly, this metric quantifies semantic agreement between post-hoc model and expert explanation texts across the complete answer set; it does not measure the latent reasoning process that produced the model's selected answer. For each answer option, the generated and reference text was segmented into sentences and embedded with a sentence-transformer model (all-mpnet-base-v2). From the resulting cross-text cosine-similarity matrix, we computed the mean similarity as F1 × (1 − contradiction), where F1 is the harmonic mean of coverage and precision, and contradiction is the mean contradiction probability from a natural-language-inference cross-encoder (nli-deberta-v3-base) over aligned sentence pairs. Coverage and precision are computed in a manner similar to the recall and precision scores of a BERTScore-style greedy sentence matching (each sentence scored against its closest counterpart in the other text). This formulation rewards rationales that semantically cover the expert explanation while penalizing those that assert claims contradicting it.

The pseudocode below provides more context on how the similarity score is computed from the expert explanation *E* and model explanation *M*:

```
1. Split E and M each into sentences and embed every sentence with a sentence-transformer (all-mpnet-base-v2, L2-normalized).
2. S[i, j] = cosine(E_i, M_j)                    # cross-text similarity matrix
3. coverage  = mean_i ( max_j S[i, j] )          # recall over expert sentences
   precision = mean_j ( max_i S[i, j] )          # precision over model sentences
   F1        = 2 · coverage · precision / (coverage + precision)
4. pairs = row-wise and column-wise argmax matches of S
   contradiction = mean over pairs of  max( NLI_contr(E_i, M_j), NLI_contr(M_j, E_i) )
5. similarity = F1 × (1 − contradiction)
```

### *V-7.* Literature prompts and similarity scoring

For literature synthesis, each model was given the abstract-withheld article text, including retained figure captions but not separate metadata such as the stored article title or DOI. To enable standardized evaluation across models with heterogeneous context-window capacities, particularly smaller locally deployable models, inputs were capped at approximately 50,000 tokens. Articles exceeding this limit were truncated to the first approximately 50,000 tokens, whereas shorter articles were provided without truncation. The model was then prompted, as a domain scientist, to write a publication-ready abstract of approximately 150–250 words summarizing the research question, methods, results, and conclusions. Each generated abstract was scored against the original author-written abstract using the same sentence-embedding similarity metric described in V-6 (coverage, precision, F1, and the NLI contradiction penalty; final score = F1 × (1 − contradiction)). Per-model scores were averaged within each subfield to produce the subfield-resolved similarity heatmap (**Fig. 4B**).

To characterize the background scale of the L-Syn similarity metric for scientific texts describing different studies, we constructed a reference distribution using mismatched pairs of the 218 author-written abstracts. We enumerated all 23,653 unique abstract pairs and scored them using the same procedure applied to model-generated abstracts (V-6: all-mpnet-base-v2 sentence embeddings, greedy coverage/precision matching, F1, and the nli-deberta-v3-base contradiction penalty). One article had been submitted twice under different research questions and was identified by an identical DOI; the corresponding duplicate pair was excluded, leaving 23,870 mismatched pairs for analysis. Pairs were stratified by topical relatedness into three tiers: different subfield (n = 20,710), same subfield but different research question (n = 2,724), and same research question (n = 436). The last tier comprised distinct articles nominated by experts under the same active research question and therefore represented the most closely related non-matching articles available in the benchmark. Because the contradiction term is defined at the sentence-pair level rather than the abstract-pair level, the sentence-level contradiction matrix was computed once across all 1,600 abstract sentences and reused across abstract pairs. The mean contradiction penalty over aligned sentence pairs was 0.162 and was therefore retained in full rather than approximated or omitted. As a positive control for the scoring procedure, the duplicated article scored 0.997 when compared with itself. These distributions were used to characterize the similarity expected between distinct scientific articles of varying topical relatedness and were not treated as a statistically matched null distribution for the model–subfield mean scores reported in **Fig. 4B**.

***V-8.* Multimodal prompts, accuracy and expert–model explanation-text similarity scoring**
The pilot set of 10 expert-curated, image-based questions was evaluated with the same two-stage protocol and scoring as the MCQ benchmark (V-6), with each question's experimental image (base64-encoded, using provider-specific formatting) attached to both the answer and rationale calls. The 16 vision-capable models were scored on answer accuracy and expert–model explanation-text similarity using the same option-wise procedure described in V-6, with the four option-level explanation similarity scores averaged for each question (**Fig. 5A** & **B**). The pilot dataset will be expanded in an upcoming version of this manuscript.

***V-9.* Correctness-pattern agreement and ensemble analysis**
To assess whether models showed similar patterns of success and failure across questions, we represented each model as a 332-dimensional binary correctness vector, with 1 indicating a correct response and 0 indicating an incorrect response. Pairwise Cohen's $\kappa$ was then computed between models to quantify agreement in these binary correctness patterns beyond chance. Because this representation records only whether an answer was correct or incorrect, $\kappa$ does not indicate whether two models selected the same incorrect answer when both failed. Note that the 27 questions (26 answered correctly by all models and 1 by none), which carry no inter-model signal, were excluded in this analysis. Models were then hierarchically clustered on the $\kappa$-distance $d = 1 - \kappa$, and the dendrogram was cut at $d = 0.5$ to identify clusters of similar correctness patterns (**Fig. 6**).

To test whether combining models improves accuracy, we performed an exhaustive majority-vote search over all 3-, 5-, and 7-model ensembles of the 21 models. For each question, the predicted option was the one receiving the most votes, with ties broken by the vote of the ensemble member with the highest overall accuracy on the same MCQ benchmark. Each ensemble was summarized by its majority-vote accuracy and its mean within-ensemble κ, where lower κ indicates greater diversity in binary correctness patterns. For each ensemble size, we identified the in-sample best ensemble as the combination achieving the highest accuracy on the same 359-question benchmark and compared it with the 'naive top-k' ensemble comprising the k individually highest-accuracy models (**Fig. 6**). Ensemble selection and evaluation were therefore performed on the same question set, without a separate held-out set.

### *V-10.* Statistical analysis and visualization

Model performance was reported as mean accuracy (the proportion of correct answers), while similarity scores were reported as distributions with the mean marked on violin plots. No inferential hypothesis tests were performed for comparisons of model performance, subfield accuracy, explanation-text similarity, or ensemble performance. The only inferential tests in this study were the two-sided Fisher's exact tests used in the post-evaluation 40-item MCQ audit described in V-4.2. Data handling and statistics used Python (pandas, NumPy, scikit-learn for Cohen's κ, and SciPy for hierarchical clustering); semantic-similarity scoring used sentence-transformers (V-6–V-7); and figures were generated with matplotlib and seaborn.

### *V-11.* Data and code availability

The data will be provided via a unified API for evaluation and model development. Still, they will not be released to the public domain to prevent accidental use of the dataset during model training. The API is expected to both serve the questions (without identifying the correct answer) and evaluate the accuracy of a model's response. The API access requests will be reviewed by the coordinating team against documented criteria, with this function transitioning to a multi-institution governance committee as the consortium expands. Code used to generate all the figures in the manuscript will also be available on GitHub prior to the final publication of the manuscript.

## VI. Acknowledgement

We thank Serendipity Lab for providing collaboration opportunities. M-T.I. was supported by MIT-Novo Nordisk Artificial I Reliance Fellowship. S.S.K. was supported by the Swiss National Science Foundation (SNSF) project 216632. H.H. was supported by the UCLA MIMG Warsaw Graduate Fellowship. S.T. was supported by NIH R35GM148231. D.D. was supported by the EXPERT-J program of the Japan Science and Technology Agency.

## VII. Conflict of Interest

K.G. is a shareholder of CYBO, LucasLand, FlyWorks, and NanoTitan.